%% file: main.tex
\documentclass[10pt,twocolumn,letterpaper]{article}

\usepackage{amsmath}
\usepackage{amssymb}
\usepackage{amsfonts}         
\usepackage{algorithmicx}
\usepackage{algorithm}
\usepackage{array}
\usepackage[noend]{algpseudocode}
\usepackage{booktabs}         
\usepackage{bm}
\usepackage{calc}
\usepackage{capt-of,etoolbox}
\usepackage{epsfig}
\usepackage{float}
\usepackage{graphicx}
\usepackage[pagebackref=true,breaklinks=true,colorlinks,bookmarks=false]{hyperref}
\usepackage{iccv}
\usepackage{listings}
\usepackage{microtype}        
\usepackage{multirow}
\usepackage{nicefrac}         
\usepackage{subcaption}
\usepackage{times}
\usepackage{tabu}
\usepackage{xfrac}

\iccvfinalcopy 

\def\iccvPaperID{****} 

\newcommand{\iidsim}{\overset{\text{iid}}{\sim}}

\ificcvfinal\fi

\newcommand{\checkColor}{black}

\newcommand{\inst}[1]{\textsuperscript{#1}}

\input{figures.tex}

\begin{document}

\title{StyleGenes: Discrete and Efficient Latent Distributions for GANs}

\author{
Evangelos Ntavelis \inst{1,2}
\and
Mohamad Shahbazi\inst{1}
\and
Iason Kastanis\inst{2}
\and
Radu Timofte\inst{1,3}
\and
Martin Danelljan\inst{1}
\and
Luc Van Gool\inst{1,4}
\and
\\
{
\vspace{-1.5mm}
\inst{1} Computer Vision Lab, ETH Zurich, CH 
\inst{2} CSEM, CH 
\inst{3} University of Würzburg, DE 
\inst{4} KU Leuven, BE
}
\\
{%
\tt\small%
{entavelis,mshahbazi,radu.timofte,martin.danelljan,vangool@vision.ee.ethz.ch}
}%
}
\maketitle
\begin{abstract}
We propose a discrete latent distribution for Generative Adversarial Networks (GANs). Instead of drawing latent vectors from a continuous prior, we sample from a finite set of learnable latents. However, a direct parametrization of such a distribution leads to an intractable linear increase in memory in order to ensure sufficient sample diversity. We address this key issue by taking inspiration from the encoding of information in biological organisms. Instead of learning a separate latent vector for each sample, we split the latent space into a set of \emph{genes}. For each gene, we train a small bank of gene \emph{variants}. Thus, by independently sampling a variant for each gene and combining them into the final latent vector, our approach can represent a vast number of unique latent samples from a compact set of learnable parameters. Interestingly, our gene-inspired latent encoding allows for new and intuitive approaches to latent-space exploration, enabling conditional sampling from our unconditionally trained model. Moreover, our approach preserves state-of-the-art photo-realism while achieving better disentanglement than the widely-used StyleMapping network. 
\end{abstract}

\input{./source/introduction}

\input{./source/relatedwork}
\input{./source/method}
\input{./source/experiments}

\input{./source/conclusion}

\noindent\textbf{Acknowledgements}
This work was partly supported by CSEM and the ETH Future Computing Laboratory (EFCL), financed by a gift from Huawei Technologies.

{\small
\bibliographystyle{ieee_fullname}
\bibliography{eglib}
}

\appendix

\input{./source/limitations}


\section{Evolution of StyleGenome during training}
\begin{figure*}
 \centering
 \begin{tabular}{c}
     \includegraphics[width=0.95\linewidth,trim={80 0 50 20}]{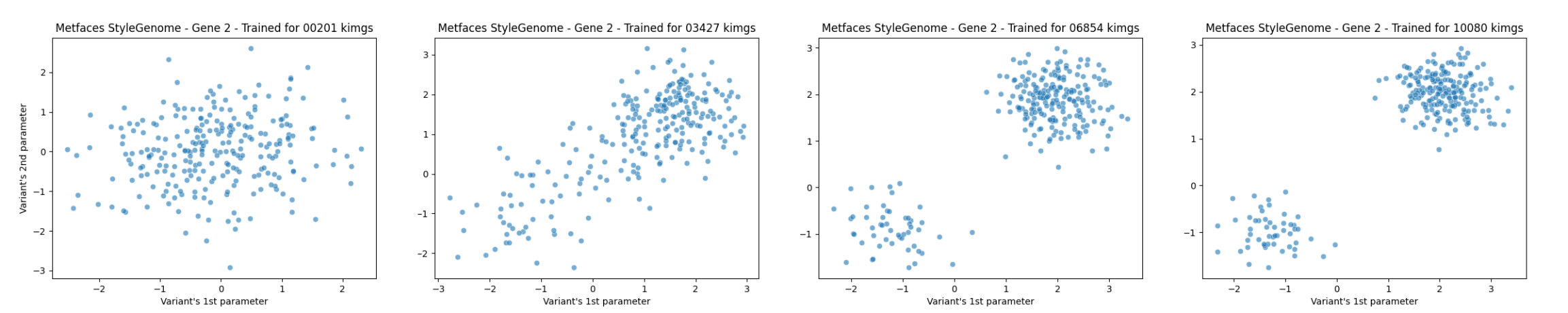} \\
 \end{tabular}
 \caption{We can observe how the density of the discrete latent distribution is changing through training. The variants' values are learned through the adversarial game to produce images, along with the synthesis network. that match the real distribution.}\vspace{-5mm}
\label{fig:genometraining}
\end{figure*}

As we discussed in Section 4 of the main paper, the StyleGAN's~\cite{Karras_2019_CVPR} motivation to design the StyleMapping Network to make the sampling density determined by the mapping and not to be limited to any fixed distribution;
they aimed for the resulting space W to be more disentangled.
In Figure~\ref{fig:genometraining} we can see how the adversarial game is altering the density of our discrete distribution throughout the training.
Our learnable embeddings are aligning, together with the Synthesis network, to better match the real images distribution.

\

\section{Pruning the genome}

There is another common technique applied to continuous latent spaces of GANs that we have not addressed: the truncation trick\cite{brock2018large}. As we have an unordered set of variants, applying the trick is not straight-forward. However we develop a technique to limit the erroneous samples that our model can synthesize.
Using the discriminator as a heuristic, we replicate the process we follow in the main paper to produce the expected value of an attribute. Similarly, we derive the expected value of \emph{realness} of a particular variant as the average discriminator output over a set of samples that contain this variant. 

As with the truncation trick, we limit the number of samples the network can generate by removing certain variants off the \emph{gene pool}.
In Table \ref{tab:pruning} we compute the FID  score for different genome scenarios. 
We remove $x$ number of variants that exhibit the lowest \emph{realness} in expected values and substituted them with the ones that exhibit the highest.
The number on the top of each column denotes this number for each scenario.
The configuration we used for this experiment has 1024 variants per gene, making the scenario of the rightmost column have half the size of the genome of the leftmost one.

We find out that our pruning approach can increase the perceptual quality of the method in terms of FID. However, we hypothesize that limiting the number of variants to a larger degree can decrease the variability of the generated samples and hurt the score. Note, that we use the discriminator's score as an easy heuristic. However, samples produced using variants with extreme fake expected values are not necessarily erroneous, nor the ones with the realest values are guaranteed to be perceptually good. However, on average they produce a better FID score as shows in Table \ref{tab:pruning}.

\begin{table}
\small
\centering
\newcolumntype{h}{>{\centering\arraybackslash\hspace{0pt}}p{0.08\linewidth}}%
\resizebox{\columnwidth}{!}{%
\begin{tabular}{|h|h|h|h|h|h|h|h|}
\hline
\multicolumn{8}{|c|}{\bf Variant Pruning - FID$\downarrow$} \\
\hline
0 & 8 & 16 & 32 & 64 & 128 & 256 & 512 \\
\hline
$ 5.87 \pm 0.06$ & $ 5.81 \pm 0.06$ &$ 5.79 \pm 0.02$ &$ 5.78 \pm 0.04$ &$ \textbf{5.76} \pm 0.05$ &$ 5.91 \pm 0.05$ &$ 6.11 \pm 0.04$ &$ 6.27 \pm 0.05$ \\
%
\hline
\end{tabular}%
}%
\caption{Using our discriminator as a heuristic, we compute the expected value of \emph{realness} for each variant. We use these values to substitute the most fake variants with duplicates of the most real ones. The number on the top of each columns shows how many variants are pruned. Our pruning approach can decrease the FID. Pruning too much, however, decreases the size of the genome and can increase the score. We report the FID computed with 50k images, averaged over five runs.}\vspace{-3mm}%
\label{tab:pruning}%
\end{table}

\section{Additional configurations}

In the main paper we show results for five different datasets for images generated at $256 \times 256$ resolution using the StyleGAN2\cite{Karras2019stylegan2,Karras2020ada} synthesis network and the FastGAN Generator of ProjectedGAN\cite{Sauer2021ProjectedGC}.

We also train our \emph{StyleGenes} approach for FFHQ images of resolution $1024 \times 1024$ for 2 million images seen by the discriminator. Similarly with our main paper experiments, our results, with an FID score of $7.60$ are on par with the mapping network's FID of $7.61$. In Figure \ref{fig:1024} we can observe results generated by our approach.

For a second experiment, we substitute the StyleGAN2 backbone with StyleGAN3-T~\cite{stylegan3Karras2021} and train for the Metfaces dataset. Again, we observe similar FID scores for our method ($27.17$) and the standard approach ($27.44$)). We trained both networks for 6 million images per discriminator. We can see the samples synthesized by StyleGenes in Figure \ref{fig:stylegan3}. 

Lastly, we also train for 3d-aware image synthesis using EG3D~\cite{eg3d} for the cats sub-dataset of AFHQ. We train the method we both its original StyleMapping approach and our StyleGenes, using the same settings. We get an FID score of $4.17$ for the baseline and $4.15$ for our approach.

\begin{figure}[t]
\scalebox{1.00}{%
\includegraphics[width=\linewidth,trim={0 0 0 0}]{supp_figures/images1024_small.pdf}} %
\caption{{\bf Unconditional Generation on FFHQ $1024 \times 1024$~\cite{Karras_2019_CVPR}} using \emph{StyleGenes} along with a StyleGan2 synthesis network. All images were produced by gene sequences selected at random and without cherry-picking. }\vspace{-3mm}
\label{fig:1024}
\end{figure}

\begin{figure}[t]
\scalebox{1.00}{%
\includegraphics[width=\linewidth,trim={0 0 0 0}]{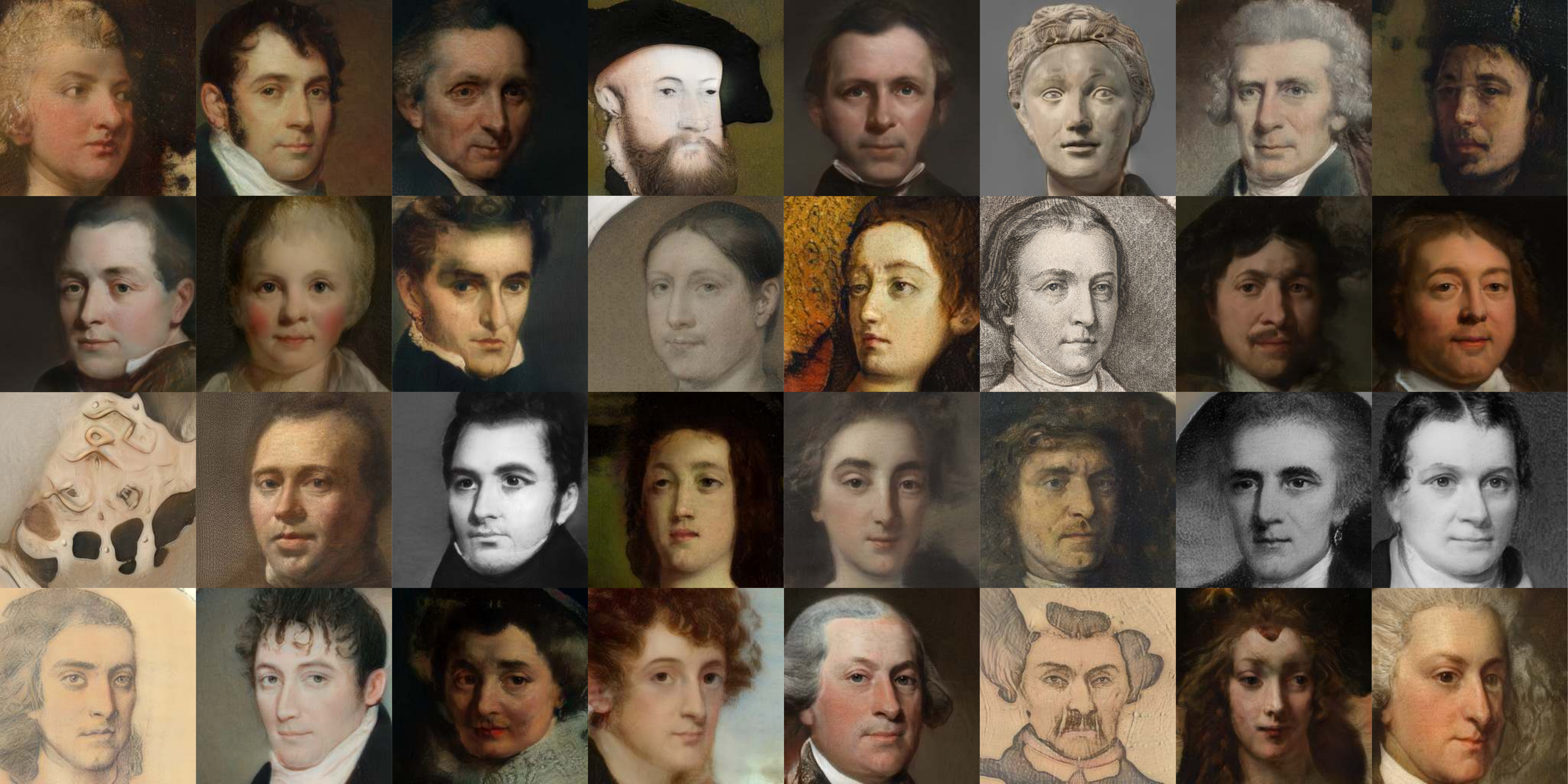}} %
\caption{{\bf Unconditional Generation on Metfaces~\cite{Karras2020ada}} using \emph{StyleGenes} along with a {\bf \emph{StyleGan3}} synthesis network. All images were produced by gene sequences selected at random and without cherry-picking.}\vspace{-3mm}
\label{fig:stylegan3}
\end{figure}

\begin{figure*}[t]
\scalebox{1.00}{%
\includegraphics[width=\linewidth,trim={0 0 0 0}]{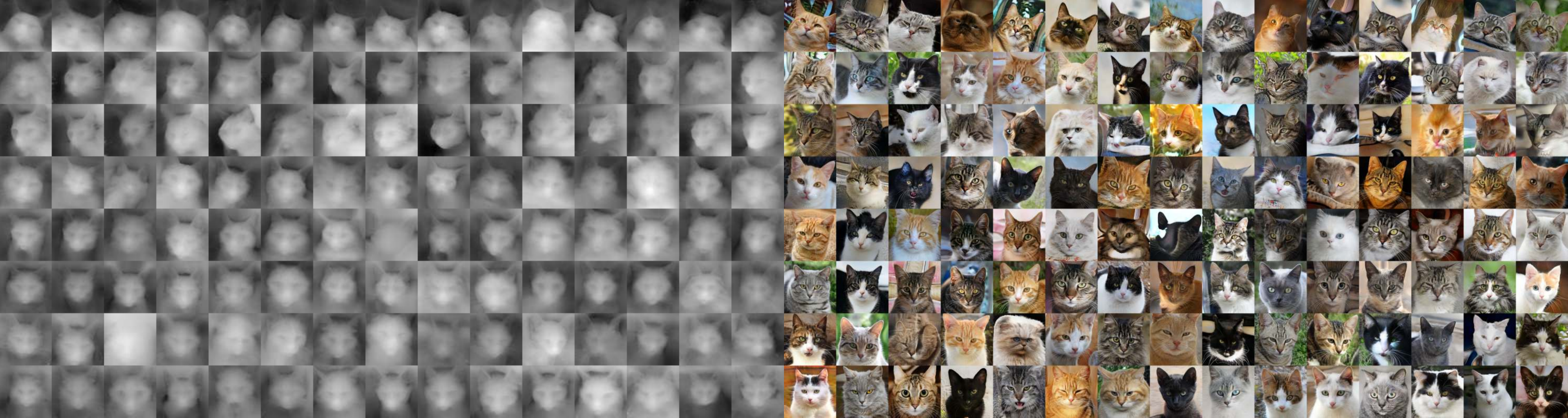}} %
\caption{{\bf 3D-aware Generation on Cats~\cite{Karras2020ada}} using \emph{StyleGenes} along with a {\bf \emph{EG3D}}~\cite{eg3d} synthesis network. All images were produced by gene sequences selected at random and without cherry-picking.}\vspace{-3mm}
\label{fig:eg3dcats}
\end{figure*}

\section{The benefits of the mapping network}

In the nominal work of the first StyleGAN~\cite{Karras_2019_CVPR}, the authors motivate the design of the mapping network as a mean to unwarp the gaussian prior in a way that permits only sampling from valid combinations of attributes. Moreover, they argue that a benefit of the newly acquired latent space is that it does not follow a predetermined distribution and it can learn its own sampling density.

Our approach also shares these benefits. Having a set of learnable discrete samples means that these \emph{move} during training, and can alter their density. While we uniformly sample the variant of each gene, training pushes these variants to be closer or further apart. Moreover, using pretrained CelebA\cite{celeba2015liu} classifiers we can quantify the correlation of certain attributes. In Figure \ref{fig:heatmaps} we see the Pearson correlation computed over the vector of classifiers' outputs for 50 thousand images. We show results for real images of the FFHQ dataset, as well as generated images from a StylegGAN using the mapping network and our approach. We can observe that the correlation values are similar between our proposed method and the mapping network. 
However, as we show in the main paper, our approach permits the sampling of latent codes that we use to predict the presence of the output with much higher accuracy than the baseline method. This indicates that our space has better disentanglement than the one from StyleMapping.

\begin{figure}[t]
\scalebox{1.00}{%
\includegraphics[width=\linewidth,trim={0 0 0 0}]{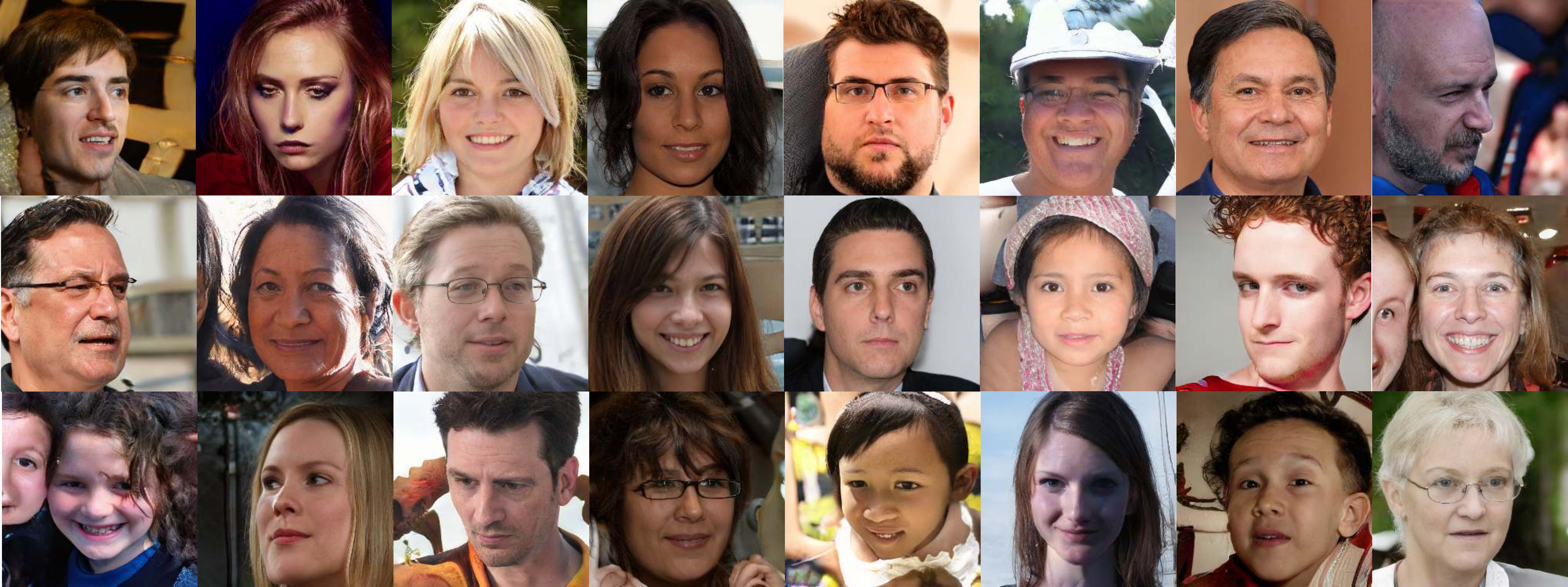}} %
\caption{{\bf Unconditional generation on FFHQ.} All images were produced by gene sequences selected at random and without cherry-picking.}\vspace{-3mm}
\label{fig:more_ffhq}
\end{figure}

\begin{figure}[t]
\scalebox{1.00}{%
\includegraphics[width=\linewidth,trim={0 0 0 0}]{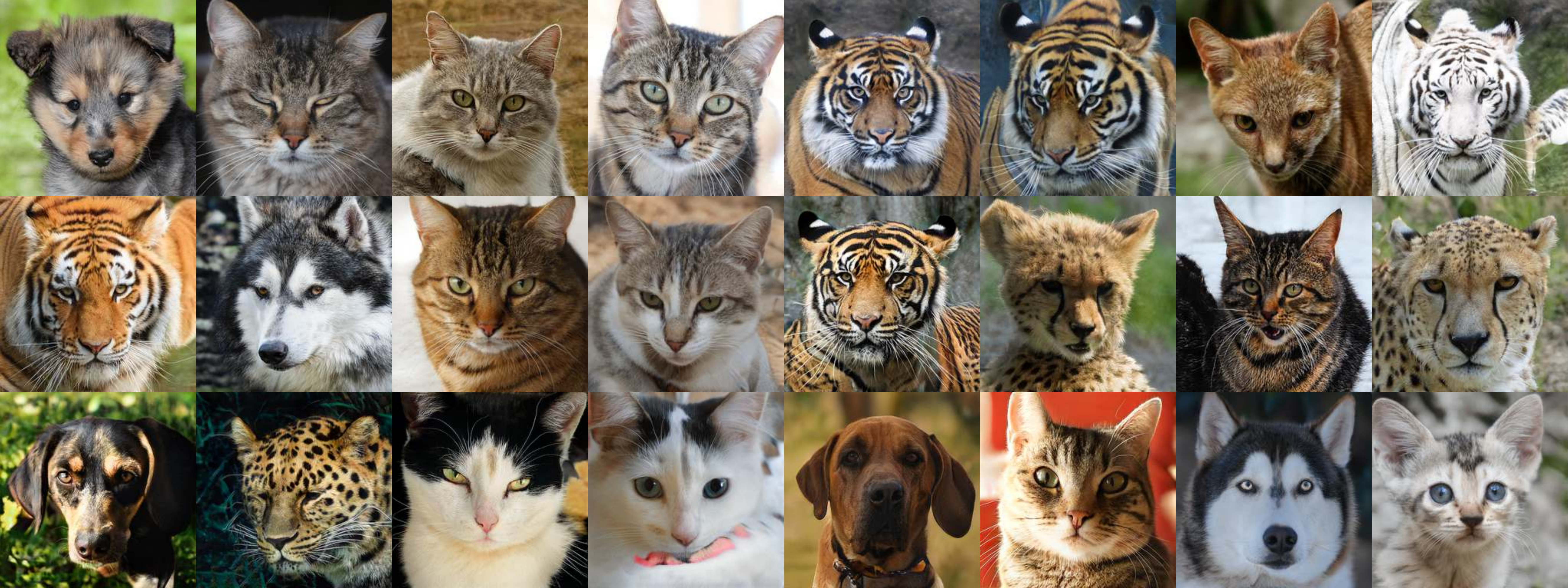}} %
\caption{{\bf Unconditional generation on AFHQv2.} All images were produced by gene sequences selected at random and without cherry-picking.}\vspace{-3mm}
\label{fig:more_afhq}
\end{figure}

\begin{figure}[t]
\scalebox{1.00}{%
\includegraphics[width=\linewidth,trim={0 0 0 0}]{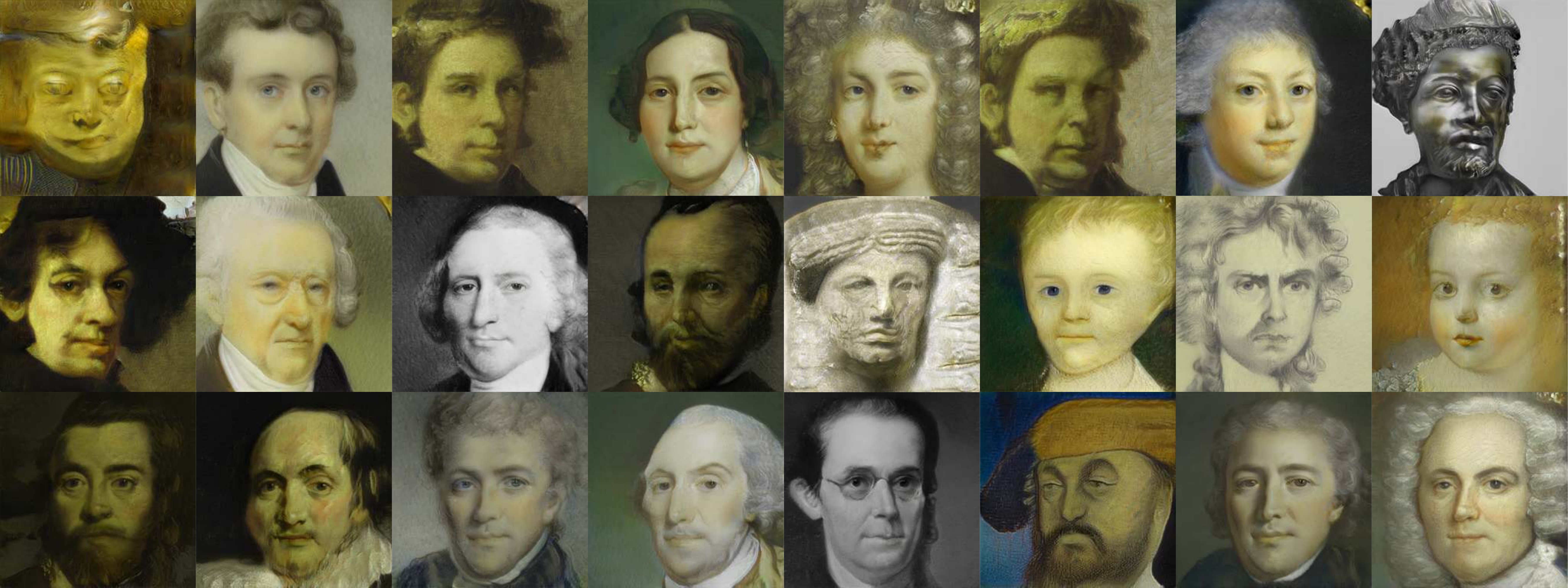}} %
\caption{{\bf Unconditional generation on Metfaces.} All images were produced by gene sequences selected at random and without cherry-picking.}\vspace{-3mm}
\label{fig:more_metfaces}
\end{figure}


\section{Projecting a real image via latent optimization}

\begin{figure}[t]
\scalebox{1.00}{%
\includegraphics[width=\linewidth,trim={0 0 0 0}]{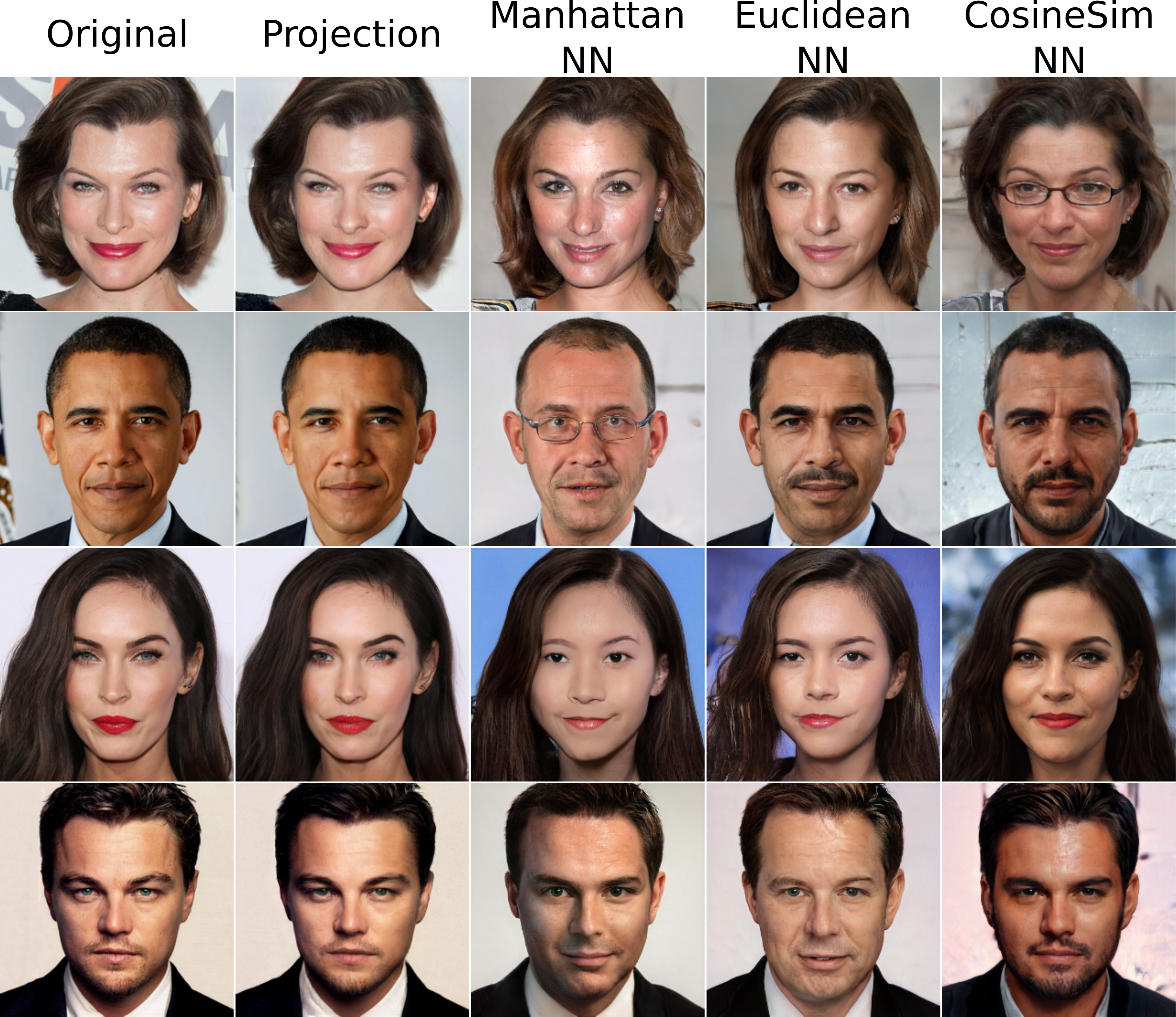}} %
\caption{{\bf Image Inversion with latent code optimization.} While projecting specific faces to our discrete latent space cannot guarantee good results. Optimizing in our underlying continuous space is able to produce \emph{projection} samples of high quality and fidelity. In the three last columns we find the closest variants of each gene to the real projected vector, using different distance functions: \emph{Manhattan} distance, \emph{Euclidean} distance and \emph{Cosine Similarity}. We use this us an intermediate step for our genome inversion, explained in the experiments section of the main paper.}\vspace{-3mm}
\label{fig:projection}
\end{figure}

In the main paper we discussed how the discrete nature of our proposed \emph{StyleGenome} enables us to conditionally generate and manipulate images. Even if the genome has a finite number of images it can generate, this number is large enough to produce countless different samples. However, it is interesting to approach the inverse problem. We have a specific image, how can we \emph{project} it to the latent space? 

In the main paper we have shown that we can generate realistic samples using real vectors in between two of our gene sequences.
Similarly, we tackle the task of projection to the latent space as a continuous optimization problem.
In contrast to StyleGAN's\cite{Karras_2019_CVPR} projection~\cite{embedstylegan}, where a set of samples from a gaussian distribution is passed through the mapping network and then averaged, we average the real values of randomly selected gene sequences.
In the first two columns of Figure \ref{fig:projection}, we can see the real images and our method's projections.
We optimize the style vectors modulating each layer of the synthesis network ($W^{+} space$)~\cite{embedstylegan}.
The vector is optimized in order to minimize mean squared error and the perceptual distance~\cite{zhang2018perceptual} between the synthesized image and the original image we want to project.

Training with our discrete set of genes we are able to learn a dense continuous space that enables projection of images to the level shown in Figure \ref{fig:projection}.
As we can see in the rest of the columns of the figure, however, the closest variants to the real sub-vectors producing the projected image, can not recreate the original well. In Figure \ref{fig:projection}, we show three different approaches for computing the distance of the projected latent code's sub-vectors to the genome's variants: Manhattan distance, euclidean, and cosine similarity. 

In practise we use this approach to initiate our Codebook inversion, as described in the experiments section of the main manuscript. 
\section{Additional visual results.}
In the following figures we provide additional results:
\begin{itemize}
\item In Figure \ref{fig:more_ffhq} we present more unconditional samples for FFHQ~\cite{Karras_2019_CVPR}. 
\item In Figure \ref{fig:more_afhq} we present more unconditional samples for AFHQv2~\cite{choi2020starganv2}. 
\item In Figure \ref{fig:more_metfaces} we present more unconditional samples for Metfaces~\cite{Karras2020ada}. 
\item In Figure \ref{fig:polars} we are presenting a clearer version of our gene analysis, shown in Figure 2 of the main paper. 
    
\end{itemize}
\section{Genome Codebase}
In Figure~\ref{fig:code} we share the sampling code of \emph{StyleGenes}. Using the StyleGAN3~\cite{stylegan3Karras2021} codebase: \url{https://github.com/NVlabs/stylegan3}, adding this function will enable training with our approach. 

\begin{figure*}[t]
\begin{lstlisting}[language=python]
# Modified from:
# github.com/NVlabs/stylegan3/
import torch
class Genome(torch.nn.Module):
    def __init__(self,
        dim,  # style latent dimensionality.
        num_ws, # number of W copies used to modulate the synthesis network
        num_variants = 2048,
        gene_length = 8,
   ):
        super().__init__()
        self.dim = dim
        self.num_ws = num_ws 

        #Discrete genome
        self.gene_length = gene_length
        assert (self.dim % self.gene_length) == 0
        self.num_genes = self.dim // self.gene_length
        self.num_variants = num_variants 
        genes = torch.randn((self.num_genes,  
                              self.num_variants, 
                              self.gene_length), 
                              requires_grad=True)
        self.genes = torch.nn.Parameter(genes)

        #
        self.gene_inds = torch.arange(self.num_genes).view(1, -1)  
        
    # We use the same arguments as the Mapping network
    # in order to seamlessly substitute the function call
    def forward(self, z, \
                      c, \
                      truncation_psi=1, \
                      truncation_cutoff=None, \
                      update_emas=False):
                      
        # We randomly select one variant per gene
        # to create the gene sequence
        
        batch_size = z.shape[0]
        gene_inds = self.gene_inds.broadcast_to(batch_size, self.num_genes)
        var_inds = torch.randint(size=(batch_size, self.num_genes),
                    high= self.num_variants)
        ws = self.genes[gene_inds, var_inds].view(batch_size, -1)

        # This is only for StyleGAN-based architectures
        # We remove it for FastGAN 
        ws = ws.unsqueeze(1).repeat([1, self.num_ws, 1])

        return ws

\end{lstlisting}
\caption{The code for StyleGenes. For StyleGAN-based architectures we use our Genome to substitute the StyleMapping Network. For the FastGAN generator we use it in the stead of the gaussian sampling}
\label{fig:code}
\end{figure*}

\begin{figure*}[t]
\scalebox{1.00}{%
\includegraphics[width=\linewidth,trim={40 40 40 40}, clip]{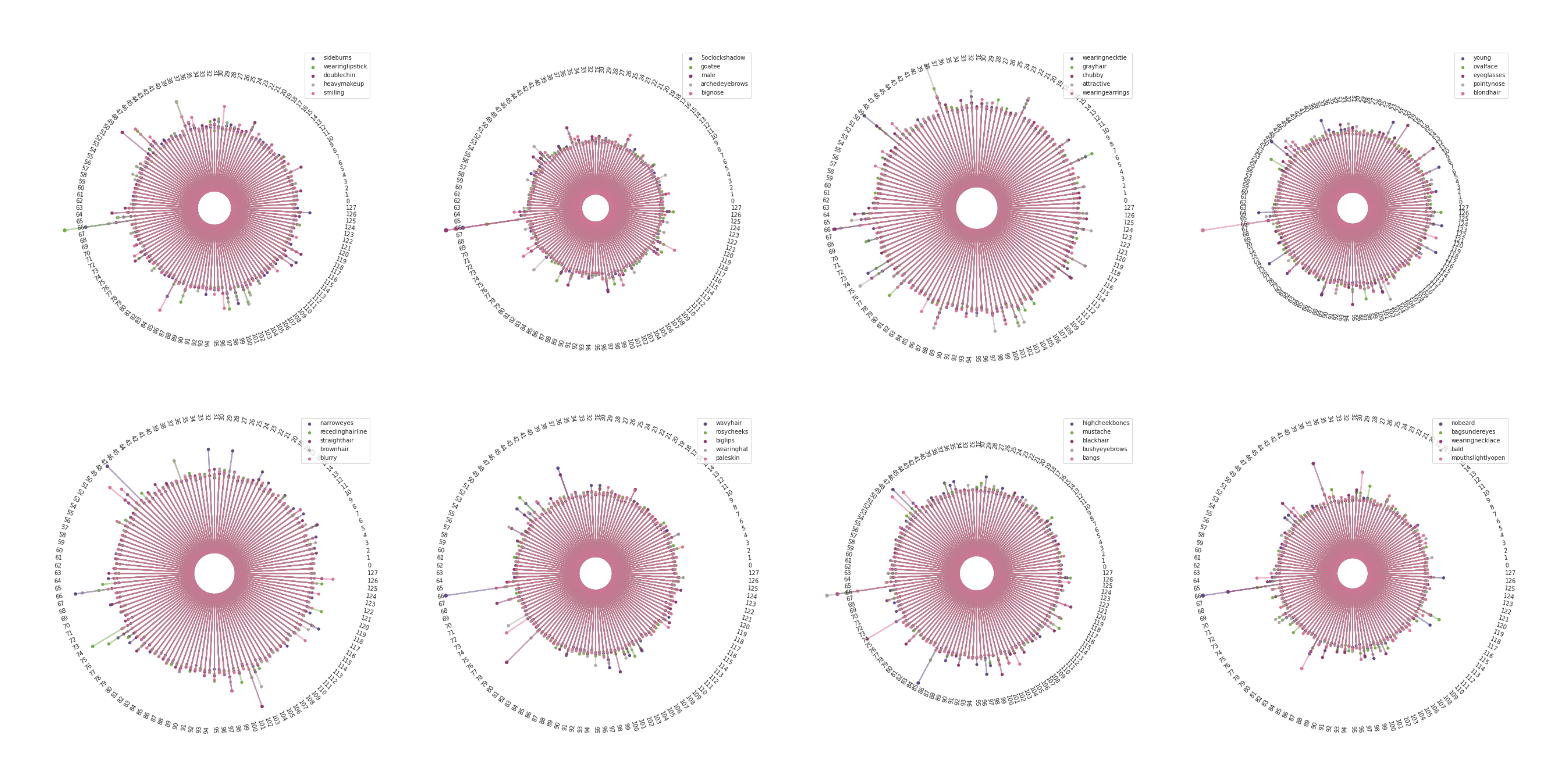}} %
\caption{{\bf Attributes' absolute standard scores.} We are presenting a clearer version of Figure 2 of the main paper.
One can easily see that for most attributes only specific genes produce high variability. These are the genes we sample more frequently in our conditional generation approach.
Most genes are only utilized for small changes in the images on their own, but combining them can create diverse outputs. Please zoom for a more detailed view.}\vspace{-3mm}
\label{fig:polars}
\end{figure*}

\begin{figure*}[t]
\scalebox{1.00}{%
\includegraphics[width=\linewidth,trim={00 00 00 00}, clip]{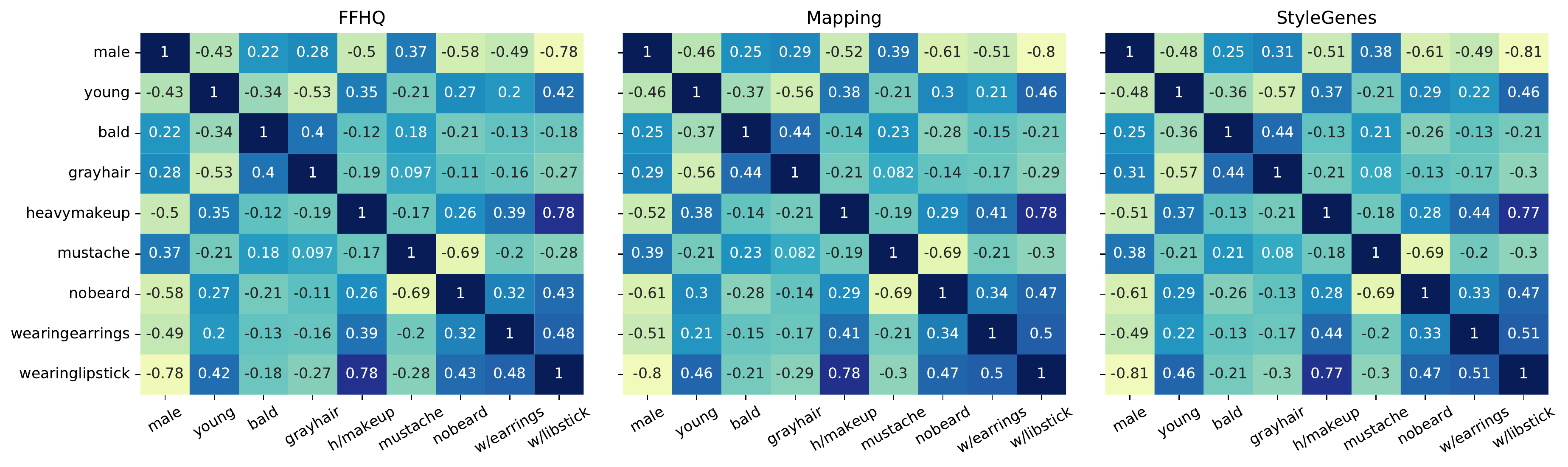}} %
\caption{{\bf Attribute Correlation.} One motivation behind the design of the mapping network is to forbid the sampling of invalid combinations, that are not present in the original dataset. By using pretrained classifiers, we show that the correlation between the certain attributes is similar between real FFHQ images and those produced by both the mapping network and our discrete sampling method.}\vspace{-3mm}
\label{fig:heatmaps}
\end{figure*}

\end{document}

%% file: figures.tex
\newcommand{\figAnalysis}[1]{
\begin{figure*}[t]
\centering
\includegraphics[width=\linewidth,trim={0 0 0 0},clip]{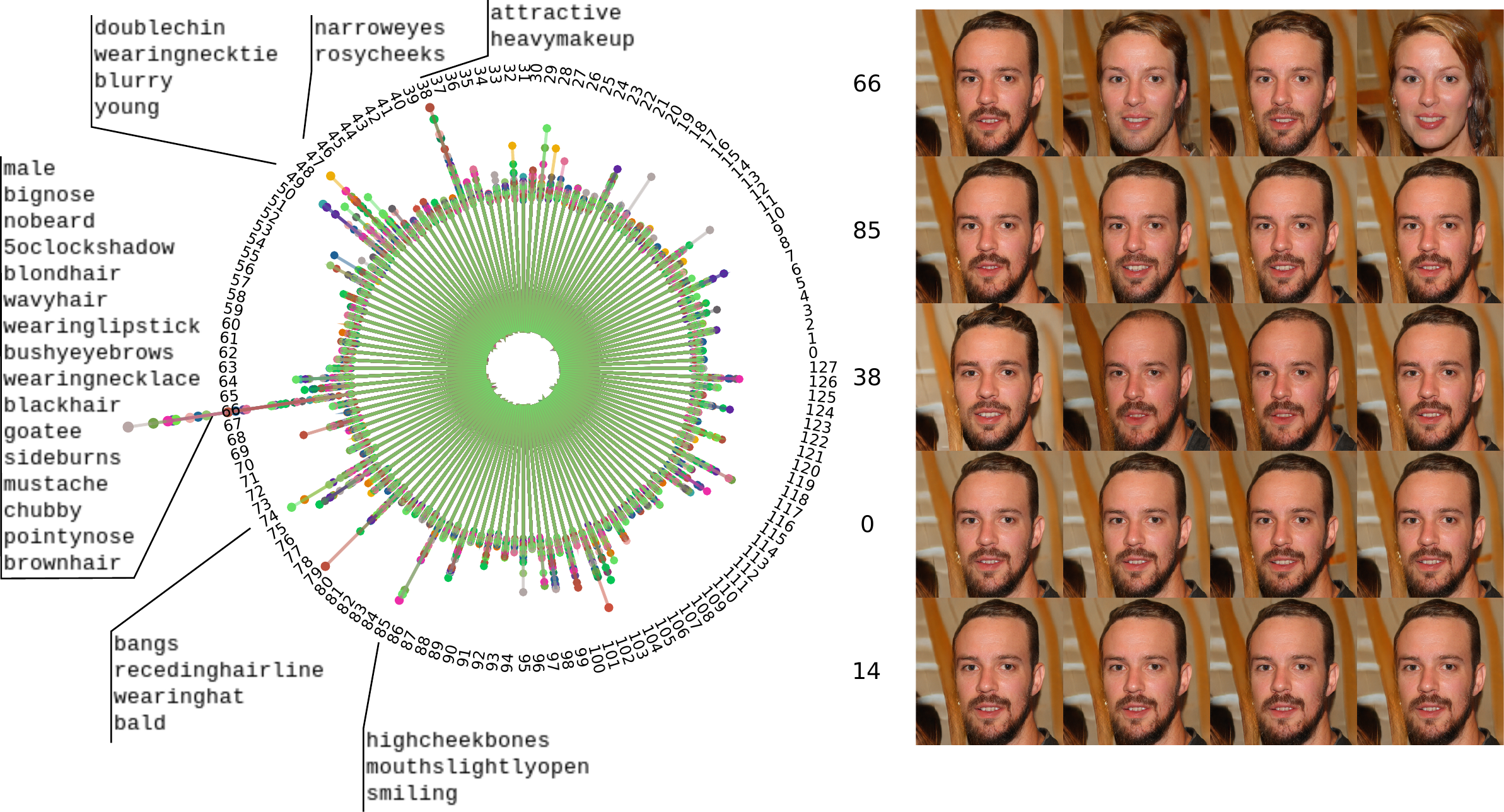}
\caption{\small{{\bf Which genes affect which attribute?} The polar plot shows which genes affect which CelebA attributes the most.
Each color represents a different attribute.
The labels indicate the gene that exhibits the highest variability in this attribute.
Most attributes are affected only by a handful of genes.
We observe that specific genes affect pertinent attributes.
For instance, genes 66 and 38 affect gender and hair color, while gene 85 the mouth-related features.
We show how randomly changing a specific gene's variants produces different alterations to the images on the right.
Changing the variants of genes 0 and 14, which do not exhibit importance for any attribute, leads to minute changes.
Most genes fall into this category.
}%
}\label{#1}%
\vspace{-3mm}%
\end{figure*}%
}

\newcommand{\figInterpolation}[1]{
\begin{figure*}[t]
\centering
\includegraphics[width=0.80\linewidth,trim={0 0 0 0}]{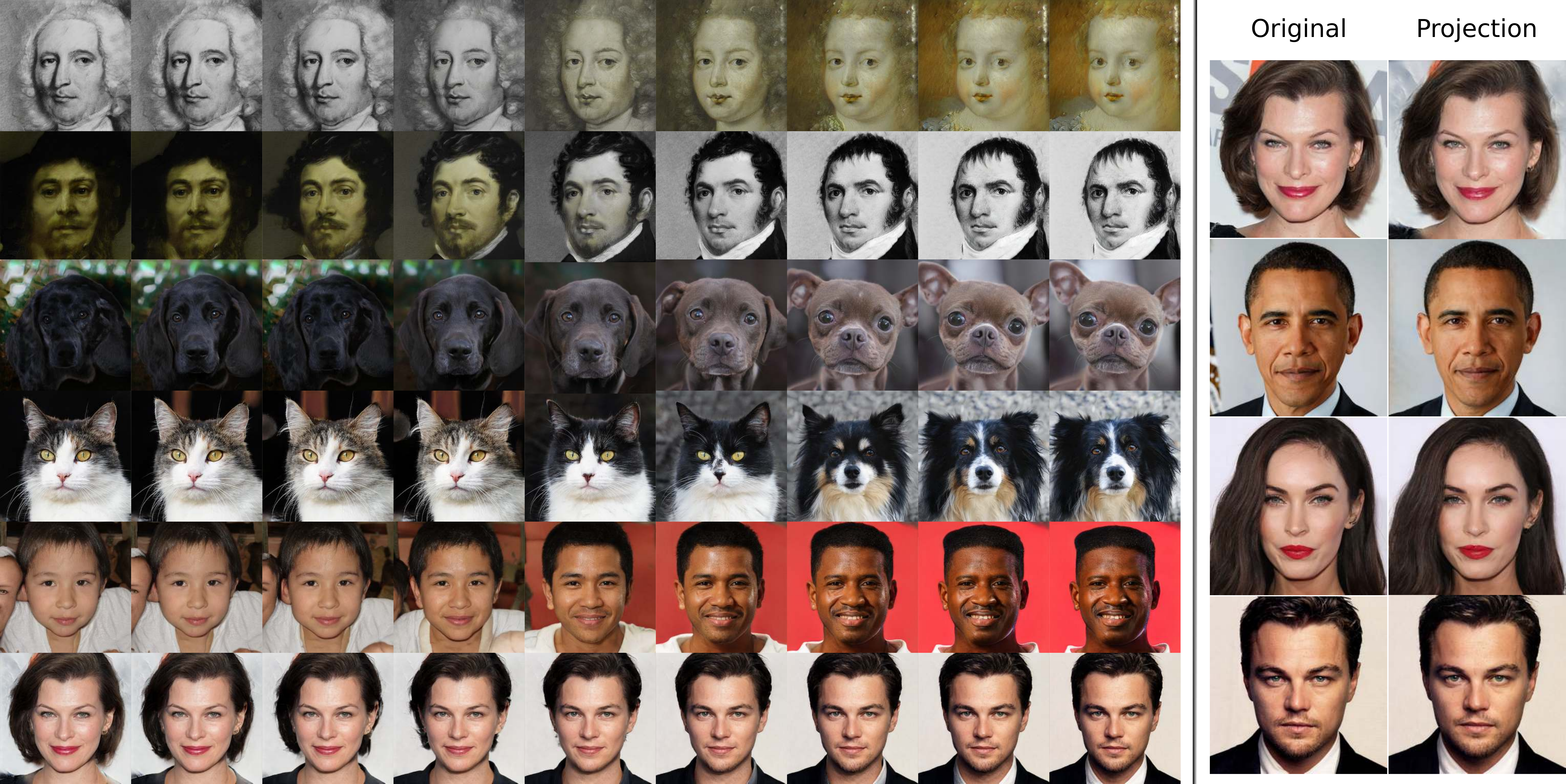}
\caption{{\bf Interpolation and Inversion.} Our latents are trained in a discrete fashion, but have real values. Thus, it is possible to interpolate between them. Our network can generate realistic results from codes outside of the genome's values. Inspired by this feature, we extend PTI~\cite{roich2021pivotal} to add real images in our genome.}
\label{#1}%
\end{figure*}
}

\newcommand{\figResultsAblation}[1]{
\begin{table*}
\hspace{0.05cm}
\begin{minipage}{.45\linewidth}
\raggedleft
\scalebox{0.70}{
\begin{tabular}{|l|ccccc|c|}
        \multicolumn{6}{c}{\large{\textbf{A: Unsupervised Image Generation}}} \\ 
\hline
      &  \multicolumn{5}{c|}{\bf FID $\downarrow$} & {\bf Time/latent $\downarrow$} \\
    \hline
  & \textbf{FFHQ} & \textbf{AFHQ} & \textbf{Met/s} & \textbf{Church} & \textbf{Beds} & \\ 
    \hline
        \multicolumn{7}{|c|}{StyleGAN2}  \\ 
    \hline
        \small{StyleMapping} & 5.3 & \textbf{5.62} & \textbf{20.48} & 8.13 & 51.61 & 0.483 ms \\ 
        \small{StyleGenes} & \textbf{5.11} & 5.99 & 21.00 & \textbf{6.86} & \textbf{17.84} & \textbf{0.170 ms} \\ 
    \hline
        \multicolumn{7}{|c|}{ProjectedGANs(FastGAN)}  \\ 
    \hline
         \small{Cont. Prior} & 5.08 & 4.02 & 15.38 & \textbf{3.05} & 3.15 & \textbf{0.015 ms} \\ 
         \small{StyleGenes} & \textbf{4.19} &\textbf{ 3.66} & \textbf{15.24} & 3.08 & \textbf{2.96} & 0.076 ms \\ 
    \hline
    \end{tabular}
}
\end{minipage}
\hspace{0.10cm}
\begin{minipage}{.4\linewidth}
\raggedright
\scalebox{0.75}{
\begin{tabular}{|c|ccc|ccc|ccc|}
        \multicolumn{10}{c}{\large{\textbf{B: Ablation Study on StyleGenome}}}\\ 
\hline
{\bf FID $\downarrow$ } & \multicolumn{3}{c|}{\bf FFHQ} & \multicolumn{3}{c|}{\bf AFHQ} & \multicolumn{3}{c|}{\bf Metfaces}          \\
\hline
\textit{Genome} & \multicolumn{3}{c|}{\# Genes} & \multicolumn{3}{c|}{\# Genes} & \multicolumn{3}{c|}{\# Genes}          \\
\hline
\small{\#Variants} & 64 & 8 & 2 & 64 & 8 & 2 & 64 & 8 & 2 \\
\hline
256 & 
5.87 & 5.34 & 24.72 & 
6.45 & 12.43 & 18.64 & 
22.56 & 38.76 & 42.60   \\
512 &
5.53 & 5.64 & 12.34 &
\textbf{5.99} & 7.24 & 13.77 & 
21.54 & 27.20 & 37.71   \\
1024 
& 5.71 & 5.20 & 6.2 
& 6.11 & 10.33 & 10.47 
& 21.99 & 25.05 & 32.08   \\
2048 &
5.22 & \textbf{5.11}  & 5.30 
& 6.31 & 6.37 & 7.31
& 21.39 & \textbf{21.00} & 30.93   \\
\hline
\end{tabular}}
\end{minipage}
\caption{%
\textbf{A:} Evaluation of our discrete sampling approach, \emph{StyleGenes}, by substituting the StyleGAN2's StyleMapping network or FastGAN's Gaussian sampling for ProjectedGAN. We achieve similar or better FID to the continuous case. 
\textbf{B:} Ablation on different configurations of the genome and our baseline.
Increasing the number of the embeddings in our codebook, the \emph{\# Variants}, increases the performance by increasing the number of parameters we are using.
We can also lower the FID by breaking the latent code into more genes of smaller lengths. This increases the number of unique codes we can sample from our genome without increasing the memory/parameters.}\vspace{-2mm}%
\label{#1}
\end{table*}
}

\newcommand{\figVectorQuantization}[1]{
\begin{table}
\centering
\scalebox{0.74}{
\centering
\begin{tabular}{|cccc|}
        \multicolumn{4}{c}{{\textbf{Comparison with Vector Quantization approaches - FFHQ - FID $\downarrow$ }}}\\ 
        \hline
        VQ-GAN & ViT-VQGAN & Ours /w StyleGAN2 & Ours /w ProjectedGAN \\
        \hline
        9.6 & 5.3 & 5.11 & {\bf 4.19} \\
        \hline
\end{tabular}%
}
\caption{ \textcolor{\checkColor}{VQ-methods are quantized and not
self-learned (different losses, encoder). They learn local
descriptors spatially aligned in a grid and require multiple
different codes for semantically rich and diverse images.
Contrary, we use a single global code, avoiding a parameter explosion with our Genome. We don’t require an
autoregressive sampler (e.g. transformer), and thus are faster.}}%
\vspace{-3mm}%
\label{#1}
\end{table}
}

\newcommand{\figDisentanglement}[1]{
\begin{table*}
\centering
\scalebox{0.80}{
\centering
\begin{tabular}{|l|ccccccccc|c|}
          \multicolumn{11}{c}{{\textbf{Predicting Attribute Presence from Latent Codes}}}\\ 
        \hline
        Method & male & young & bald & gray-hair & h-makeup & mustache & no-beard & w-earrings & w-lipstick & mean \\
        \hline
        StyleMapping & 49.74\% & 63.43\% & 96.02\% & 92.49\% & 87.71\% & 95.37\% & 79.06\% & 84.73\% & 69.30\% & 79.76\% \\ 
        StyleGenes & \textbf{86.71\%} & \textbf{82.90\%} & \textbf{97.01\% }&\textbf{ 93.68\% }& \textbf{92.09\% }&\textbf{ 95.82\%} & \textbf{91.53\%} & \textbf{86.40\%} & \textbf{85.89\% }& \textbf{90.23\% }\\ \hline
\end{tabular}%
}
\caption{We measure disentanglement by our ability to predict an attribute's presence in a generated image from its latent code. StyleGenes' codes are much easier to associate to attributes than the StyleMapping's ones.}%
\vspace{-3mm}%
\label{#1}
\end{table*}
}

\newcommand{\figConditional}[1]{
\begin{table*}[t]
\centering
\scalebox{0.85}{%
 \includegraphics[width=1.\linewidth,trim={0 0 0 0}]{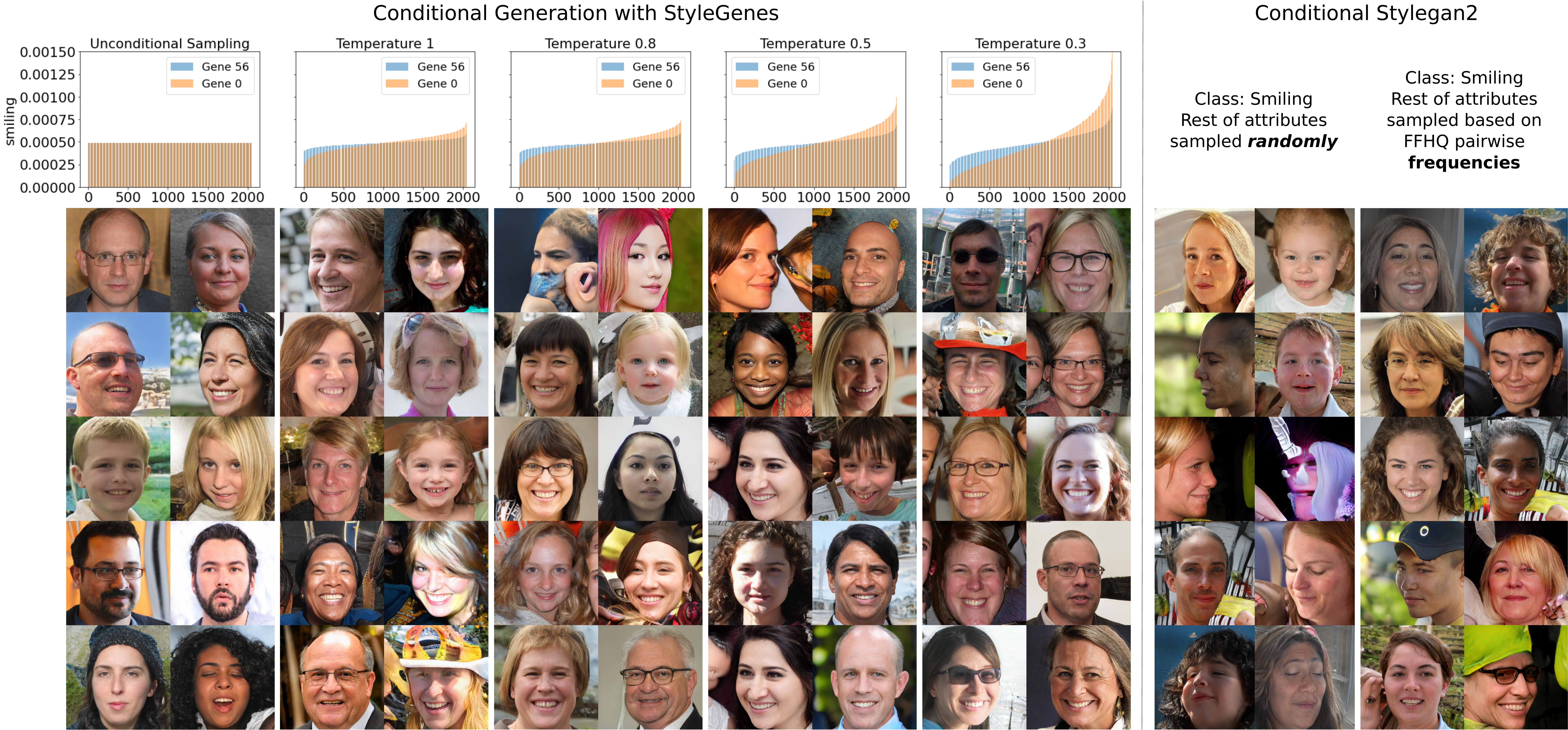}}%

\resizebox{\linewidth}{!}{
\begin{tabular}{|c|c|c|c|c|c|c|c|c|c|c|c|c|c|c|c|c|c|c|c|c|c|c|c|}
    \hline
    & & &\multicolumn{2}{|c|}{male}
    & \multicolumn{2}{c|}{eye-bags}
     & \multicolumn{2}{c|}{h-cheek/s}
    & \multicolumn{2}{c|}{smiling}
    & \multicolumn{2}{c|}{big-nose}
    & \multicolumn{2}{c|}{open-mouth}
    & \multicolumn{2}{c|}{young}
    & \multicolumn{2}{c|}{w-lipstick}
    & \multicolumn{2}{c|}{attractive}
& \multicolumn{2}{c|}{eyeglasses} \\
    Method & Sampling & Avg. & yes & no & yes & no & yes & no & yes & no & yes & no & yes & no & yes & no & yes & no & yes & no & yes & no \\ 
    \hline
    ~ & ~ &  \multicolumn{21}{c|}{Classification accuracy (\%) $\uparrow$}\\ \hline
    baseline & random & 74.54 & 93.32 & 54.24 & 70.62 & 87.66 & 32.78 & 96.46 & 35.20 & 97.76 & 59.02 & 85.74 & 83.24 & 93.16 & 91.44 & 51.34 & 44.50 & 88.52 & 29.46 & \textbf{98.82 }& 97.92 & 99.54 \\ 
    baseline & freq & 91.24 & 94.24 & 92.18 & 86.86 & 87.74 & 91.06 & 89.04 & 93.50 & 91.66 & 86.42 & 87.50 & 95.90 & 94.30 & \textbf{91.80} & 82.26 & 87.98 & 95.50 & 85.74 & 93.50 & 97.72 & \textbf{99.82} \\ 
    Ours & temp-1.0 & 71.01 & 74.26 & 74.22 & 66.28 & 85.14 & 69.82 & 75.28 & 75.70 & 74.72 & 71.96 & 69.40 & 76.58 & 71.98 & 0.38 & 65.24 & 63.88 & 81.16 & 71.50 & 96.72 & 68.76 & 87.34 \\ 
    Ours & temp-0.8 & 77.48 & 79.84 & 80.92 & 69.80 & 71.98 & 74.36 & 78.40 & 79.96 & 78.98 & 76.26 & 74.12 & 81.54 & 77.90 & 80.20 & 70.42 & 70.02 & 84.42 & 67.78 & 84.18 & 79.06 & 89.36 \\ 
    Ours & temp-0.5 & 88.02 & 90.58 & 88.36 & 80.76 & 83.28 & 86.26 & 88.00 & 90.46 & 90.84 & 86.66 & 85.20 & 91.38 & 88.36 & 87.12 & 85.40 & 85.84 & 89.58 & 84.00 & 89.54 & 96.10 & 92.70 \\ 
    Ours & temp-0.3 & \textbf{95.77} & \textbf{98.30} & \textbf{96.48 }& \textbf{91.00 }& \textbf{93.10 }& \textbf{96.26} & \textbf{96.80} & \textbf{97.82} & \textbf{97.90} & \textbf{96.00} & \textbf{95.36} & \textbf{98.14} & \textbf{93.64} & 89.62 &\textbf{ 97.48 }& \textbf{95.40} & \textbf{96.14 }& \textbf{94.56} & 95.42 &\textbf{ 99.38} & 96.52 \\ \hline
    \multicolumn{2}{|c|}{~}  &  \multicolumn{21}{c|}{FID $\downarrow$}\\ \hline
    baseline & random & 33.33 & 30.05 & 41.48 & 29.39 & 36.21 & 40.18 & 30.44 & 35.21 & 28.12 & 33.76 & 32.95 & 36.61 & 29.17 & 28.69 & 40.07 & 46.13 & 34.04 & 33.90 & 30.76 & 19.43 & 30.00 \\ 
    baseline & freq & 10.96 & \textbf{10.11} & \textbf{10.24} & 10.48 & 11.04 & 10.72 & 11.65 & 10.91 & 11.66 & \textbf{10.08} & 10.45 & 10.95 & 11.45 & 11.26 & 10.29 & \textbf{10.70} & 10.32 & 15.40 & 10.43 & \textbf{10.00 }& 11.08 \\ 
    Ours & temp-1.0 & 11.08 & 11.11 & 11.24 & \textbf{9.98} & 11.81 & 9.63 & 9.78 & 9.96 & \textbf{9.69} & 10.88 & 9.60 & 9.94 & \textbf{9.30} & 16.86 & \textbf{9.42} & 11.99 &\textbf{ 9.59} & 11.20 & 17.86 & 12.16 & \textbf{9.53} \\ 
    Ours & temp-0.8 & \textbf{10.11} & 11.04 & 10.85 & \textbf{9.88} & \textbf{10.04} & \textbf{9.60} & \textbf{9.75} & \textbf{9.67} & 9.81 & 10.72 & \textbf{9.49} & \textbf{9.89} & 9.50 & \textbf{10.17} & 9.77 & 11.43 & 9.73 &\textbf{ 10.39} & \textbf{9.70} & 11.15 & 9.66 \\ 
    Ours & temp-0.5 & 12.36 & 14.41 & 13.17 & 11.86 & 11.37 & 10.24 & 10.67 & 10.46 & 11.64 & 13.59 & 13.06 & 10.28 & 10.52 & 11.83 & 14.73 & 16.31 & 11.18 & 15.91 & 10.73 & 15.34 & 9.92 \\ 
    Ours & temp-0.3 & 28.42 & 30.92 & 28.53 & 19.43 & 17.50 & 14.97 & 14.35 & 14.24 & 79.46 & 23.46 & 28.90 & 12.99 & 14.40 & 86.22 & 32.29 & 38.63 & 18.33 & 36.92 & 15.27 & 29.86 & 11.86 \\ \hline
\end{tabular}}
\caption{{\bf Conditional Generation} We train our method unconditionally, by sampling uniformly the variants for each gene position. With our analysis we can conditionally sample the variants to generate a desired attribute, without retraining our unconditional model. We can control a FID-accuracy trade-off using the temperature. Lower values decrease variability but increase accuracy. In contrast to our method, the conditional StyleGAN baseline is limited to the attributes it was trained with. For inference they need to provide values for every attribute, and thus, to generate an image with a specific attribute, we sample the rest \emph{randomly} or use the real dataset's conditional \emph{frequencies}.} 
\vspace{-3mm}%
\label{#1}
\end{table*}
}

%% file: source/introduction.tex
\section{Introduction}


Generative adversarial networks (GANs) have seen tremendous progress since the seminal work by Goodfellow et.\ al~\cite{goodfellowGAN}.
GANs have been successfully applied to a plethora of tasks, including conditional generation from semantic categories~\cite{brock2018large, shahbazi2021cGANTransfer, shahbazi2022collapse}, images~\cite{choi2020starganv2, wang2018pix2pixHD}, text~\cite{pmlr-v48-reed16,attngan,patashnik2021styleclip}, and semantic layouts~\cite{park2019SPADE,Zhu2020SEANIS,ntavelis2020sesame,oasis}.
Compared to their early predecessors, recent GANs~\cite{stylegan3Karras2021, Sauer2021ARXIV, ntavelis2022arbitrary, chanmonteiro2020pi-GAN} are significantly more capable of realistic and diverse generation of images, with a vast number of works aimed at designing better architectures, training objectives and training strategies~\cite{karras2018progressive, Karras_2019_CVPR, Karras2020ada, liu2021towards, NIPS2017_892c3b1c}.

The core GAN formulation, however, remained largely the same: a generator transforms a latent code \emph{sampled} 
\begin{figure}[H]
\vspace{-10mm}%
\includegraphics[width=1.\linewidth]{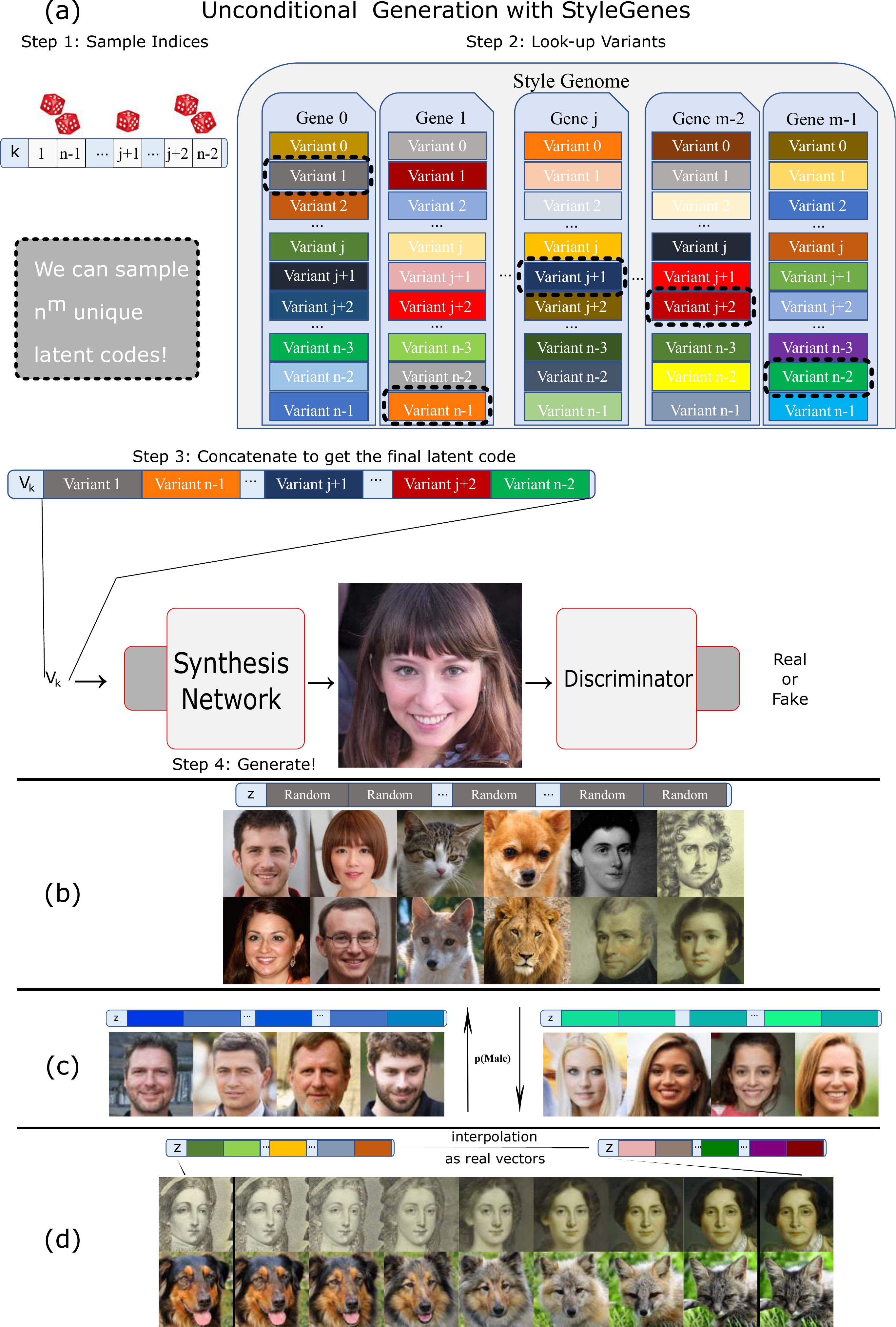}
\caption{\small{We propose \emph{StyleGenes}: a biologically inspired discrete latent distribution for GANs. Our \emph{Genome} (a) is an ordered set of smaller codebooks, we call \emph{genes}. Each gene contains a collection of embeddings, its \emph{variants}. For each gene, we select a variant and concatenate them to produce our latent code. 
We train for and perform unconditional image synthesis (b) by randomly sampling a \emph{variant} for each \emph{gene}. 
Analyzing our discrete \emph{Genome} let us associate genes with specific attributes.
We leverage this information to conditionally generate from our unconditionally trained model, without retraining or training any additional modules (c).
Although our latent distribution is discrete, the learned style space offers emergent continuous properties, ensuring smooth interpolation between samples (d).}}%
\label{fig:teaser}
\end{figure}

\emph{from a continuous distribution} to a realistic-looking image.
Initially, the latent code was sampled by a uniform distribution~\cite{goodfellowGAN}.
Quickly, however, the community converged to using a Gaussian prior~\cite{samplingGANs, improvedwgans}.
An important change came subsequently when Karras~\textit{et al}~\cite{Karras_2019_CVPR} altered the standard design of the generator network.  
The sampled noise is no longer given to the network as the initial input, but 
akin to a conditional~\cite{park2019SPADE} or a style transfer~\cite{adain2017huang} generator, it was used to manipulate the intermediate feature maps after the convolutions. Nevertheless, the 
Gaussian input is mapped to an intermediate latent \emph{style space} through a multi-layer perceptron. The motivation was that this learned space does not have to adhere to a sampling density of a fixed distribution and can be disentangled. 

Interpretation and manipulation of the GANs input space, or the latent style space, has been a subject of extensive research~\cite{jahanian2020steerability,yang2019semantic,voynov2020unsupervised,stylespaceWu2021,explaining_in_style}. 
These works usually train a separate model to make sense of the latent vector space: training a conditional normalizing flow~\cite{styleflow} or a classifier~\cite{styleEMB_neurips2021} to enable conditional sampling. For generated sample manipulation, they train one vector per transformation~\cite{jahanian2020steerability}, discover style channels via gradient computation \cite{stylespaceWu2021} or apply clustering to hidden layers~\cite{Collins2020EditingIS}. 
The use of these intricate techniques and the importance of the downstream task they are trying to tackle, raise the question on whether we can design a latent space that would permit a straightforward analysis. 



In this work, we take a different approach to continuous sampling and modulate the generator with latent codes sampled from a discrete prior distribution. 
We set the different outcomes of this distribution to be learnable embeddings, 
which induces the benefit of direct optimization of the samples.
A standard approach to designing such a discrete distribution of embeddings would require a memory bank of all the latent vectors. 
However, the advantage of an image synthesis network, is that it can generate countless novel samples.
This is not feasible with such a formulation. 

To tackle this key issue, we introduce a compact representation of a discrete distribution capable of generating an exponentially large number of distinct samples.
We draw inspiration from how the blueprint of a complex living organism, the DNA, can represent the great amount of diversity found in nature.
Only four letters, the nucleotides, form the words, the genes, that tell the story of our biology. 
A virtually endless degree of variation can be obtained by combining different variants of these genes.
Accordingly, we design our latent genome.
We break the latent code into smaller parts, the \emph{genes}. Each gene is sampled from a smaller set of gene \emph{variants}. These combine into the final latent vector, analogous to the chromosome in organisms.

We introduce \textit{StyleGenes}: an ordered collection of gene variants that are learned in conjunction with the generator, and can generate a great diversity of realistic-looking images. The nature of our latest space offers a straightforward way to interpret the associations between the discrete samples and the synthesized images. These associations can be exploited to enable downstream tasks, such as conditional generation. 

Our contributions are summarized as follows.\vspace{-2mm}
\begin{itemize}
\setlength\itemsep{0em}
\item We introduce a compact parameterization of a discrete latent distribution for GANs, inspired by the encoding of information in biological organisms.
\item
Our discrete latent space formulation permits a natural and straightforward analysis of the association of genes to semantic image attributes.
\item We use a pretrained classifier to integrate class-conditioning \emph{after} training.
Our analysis allows us to conditionally sample from the unconditionally trained model without the need to retrain, or train additional modules.
\item The learned discrete latent space is more disentangled than the widely-used StyleGAN's W space.
\item We show that despite the discrete latent distribution, the resulting style space obtains continuous properties, important for e.g.\ realistic interpolation and propose a method to project real images in our codebook.  
\end{itemize}

We perform experiments on a variety of widely-used image generation datasets and two established GAN baselines.
Our approach obtains visual results on par with the baseline continuous case, while benefiting from the intuitive gene-based approach to conditional generation  manipulation offered by our StyleGenes representation. Furthermore, our approach eliminates the need of a Style Mapping network, as it can be trained using less parameters while being \emph{faster} and yielding a more disentangled latent space.

%% file: source/relatedwork.tex
\section{Related Work}

\noindent\textbf{Latent Code Quantization:}
VQ-VAE~\cite{VQVAEOord2017NeuralDR} is one of the first studies to exploit discrete representations for image generation.
VQ-VAE is designed to prevent the posterior collapse in VAE framework when the latent representations are paired with a powerful decoder~\cite{VQVAEOord2017NeuralDR}.
Instead of a continuous latent space, VQ-VAE represents the latent space as a spatial grid of quantized local latent codes, which are sampled from a discrete set of learned vectors in an auto-regressive manner.
VQ-VAE2~\cite{VQVAE2Razavi2019GeneratingDH} is an improved version of VQ-VAE, which is capable of generating images of higher diversity and resolution by using a hierarchical multi-scale latent maps. 
The idea of VQ-VAE later on was extended to a GAN framework by changing the reconstruction loss and adding an adversarial one~\cite{VQGANEsser2021TamingTF}. Moreover, a transformer is used to learn the auto-regressive priors for sampling the discrete local latent vector.
Building on the previous approaches, RQ-VAE~\cite{lee2022autoregressive} proposes a residual feature quantization framework, which enables their model to work with smaller number of representation vectors.
Feature quantization has also been used in the discriminator of GANs to increase the stability of the adversarial training~\cite{pmlr-v119-zhao20d}.
This study bears similarities to the above works in formulating the latent space as a composition of discrete feature vectors.
However, different to prior studies, we investigate discrete sampling of the latent code in the unsupervised GAN framework~\cite{Karras2019stylegan2}, without employing any encoder or self-supervised objective. These approaches deploy an auto-encoder based approach, that produce local discrete codes and need auto-regressive sampling to draw new samples. 
Our codebook is not trained through vector quantization, but rather through the adversarial game; it provides a global description of the image to be generated and thus does not require auto-regressive sampling.

\noindent\textbf{Latent code as a composition of smaller parts:}
InfoGAN~\cite{10.5555/3157096.3157340} aimed at bringing more interpretability and disentanglement to the latent codes of GANs by maximizing the mutual information between parts of the latent code and the corresponding generated images. Inspired by the formation of DNA from genes, DNA-GAN~\cite{dnaganxiao2018} and ELEGANT~\cite{ElegantXiao2018} also proposed dividing the latent code into smaller attribute-relevant and attribute-irrelevant parts, which are then supervised using attribute annotations to create attribute disentanglement in GANs. Similar to these studies, or method divides the style vectors into smaller codes. Additionally, the style codes in our method consist of smaller codes. However, different to InfoGAN, our method only uses a discrete set of codes to form the latent style codes. Note, we do not explicitly train our method for disentanglement and feature transfer but only for unconditional image synthesis.

\noindent\textbf{Analyzing the style space:}
Steering the latent space of GANs is of high interest for many applications of image editing and conditional generation~\cite{jahanian2020steerability, yang2019semantic, voynov2020unsupervised}. Previous studies' focus has primarily been on analyzing the style space, as it is more well-behaved and disentangled compared to the traditional latent space in prior GAN models. One goal of this style space analysis is to discover meaningful directions in the style space for semantic editing of images~\cite{stylespaceWu2021, jahanian2020steerability}. Moreover,~\cite{explaining_in_style} uses style space to explaining and interpret the decisions made by attribute classifiers. The style space has also provided the opportunity for paired data generation using only a few annotations~\cite{zhang21}.
Recent methods utilize unconditionally pretrained models for conditional generation~\cite{styleflow, styleEMB_neurips2021}. These approaches, train a conditional normalizing flow~\cite{styleflow} or a classifier~\cite{styleEMB_neurips2021} in the latent space 
to enable conditional sampling.
In this study, we do not need to train one vector per transformation~\cite{jahanian2020steerability}, compute any gradients\cite{stylespaceWu2021} or apply clustering to hidden layers~\cite{Collins2020EditingIS}. In contrast, we treat the network as a black box and, without extra training, only harness the benefits of its discrete input to enable conditional generation.  

%% file: source/method.tex
\section{Method}
\label{sec:method}

In the present widely-established~\cite{Karras2019stylegan2, karras2018progressive} image generation paradigm, a latent vector sampled from a \emph{continuous} multi-variate prior distribution~\cite{goodfellowGAN} is transformed through a generator network in order to achieve the final image. In this work, we aim to offer a different approach, by starting from a \emph{discrete} distribution. We propose to sample a set of smaller latent codes from a codebook, consisting of a collection of embeddings that are trained through the adversarial learning. 




However, composing the codebook as a collection of final latent vectors leads to an intractable memory cost, as we require the generation of at least millions of unique examples.
We therefore take inspiration of how biological organisms encode information as a sequence of discrete entities, called \emph{genes}. Analogous to genes, we partition our latent vector and codebook into a sequence of \emph{positions}. At each position, we independently sample from the set of embedding \emph{variants} contained in the codebook, as illustrated in Figure \ref{fig:teaser}.a. Even with a very compact codebook, our discrete latent sampling allows for countless combinations due to the combinatorial formulation.


\subsection{Generator with continuous prior}

In the classic unsupervised image synthesis literature, the generator is a function that transforms the input noise to the image domain as,
\begin{align}
I = G(z_c) \theta_G \,, \quad z_c \iidsim p_z \,, \quad z_c \in \mathbb{R}^d
\end{align}
where $z_c$ is sampled from a prior distribution $p_z$, and $\theta$ are the generator's weights.
Early works~\cite{goodfellowGAN, Radford2015UnsupervisedRL} sample $z_c$ from a uniform distribution.
Subsequent works \cite{samplingGANs,improvedwgans} sample from a standard Gaussian distribution.
Since the introduction of StyleGAN~\cite{Karras_2019_CVPR} and the models based on it, an additional model element is deployed: a Multi-Layer Perceptron.
The weights of this \emph{mapping} network, are learned in tandem with the generator's through the adversarial objective. 
It is used as a push-forward operator to transform the Gaussian input distribution to an intermediate latent space $\mathbb{W}$.
\begin{align}
w = \text{Mapping}(z_c, \theta)\,,  \quad z_c \iidsim \mathbb{N}(0,I)
\end{align}
%
We propose an alternative method for learning a disentangled latent space $\mathbb{W}$, presented next.

\subsection{A scalable codebook of learned latent codes}

We aim to learn a discrete latent distribution.
To this end, we first introduce a codebook of $n$ learnable embeddings.
Before training, the embeddings are initialized using a standard Gaussian distribution. 
Through adversarial learning the embeddings are optimized, and therefore capable of representing flexible and complex style distributions. 
While such a formulation permits learning a set of latent codes that can generate realistic outputs, it has a fundamental flaw. The number of distinct samples we can generate scales linearly with the number of embeddings. For a latent code of length $d=512$, we would need to learn over $35$ million parameters only to be able to generate $70,000$ distinct images (the size of the FFHQ dataset~\cite{Karras_2019_CVPR}).

Inspired by how DNA encodes information in a discrete and compositional manner, we instead let the latent code be composed of an ordered set of positions, analogous to genes. 
At each position we independently and uniformly sample one of its embedding \emph{variants} from the codebook.
Then we concatenate this sequence of sampled variants into the latent code, which is used as input to the generator,
%
\begin{align}
V_k = [v_1^{k_1}, v_2^{k_2}, ..., v_{n_g}^{k_{n_g}}]\,, \quad 
{k}_i \in \{1, 2, \ldots, n_{v}\} 
\end{align}
Here, $v_i^j$ denotes the variant~$j$ of position~$i$. The vector $k$ of uniformly sampled indices $k_i$ selects the variant~$v_i^{k_i}$ for each position~$i$.
The dimensionality of $k$ is the number of positions~{s}, $n_g$, in our codebook.
The number of variants for each position is denoted $n_{v}$.
%
The final image is achieved by decoding our style vector with the generator network $G$,
\begin{equation}
I_{z_d} = G(V_k; \theta) \,.
\end{equation}
%

We let all embedding variants have the same length, such that $v_i^j \in \mathrm{R}^{d_g}$, where $d_g = d / n_g$ and $d$ is the total number of elements in the resulting latent code $V_k$.
This formulation permits the increase of distinct samples $n_{img}$ we can generate to,
\begin{align}\label{eq:combo}
n_{img} = n_{v}^{n_g} \,.
\end{align} 
For example, using a latent dimension of $d=512$ with $n_g = 64$ genes and $n_{v}=256$ variants, we can generate approximately $1.34 \times 10^{154}$ different samples; more than the estimated number of atoms in the observable universe. On the other hand, the non-compositional discrete approach using the same codebook size can only generate 256 distinct samples. In fact, by keeping $d=512$ constant, the number of trainable parameters remains independent of the number of positions $n_g$, while allowing an exponential increase in the number of distinct samples according to \eqref{eq:combo}.

\textcolor{\checkColor}{Note that our Genome is trained from scratch together with the synthesis network, guided only by the adversarial loss (See Fig.\ref{fig:teaser}).}

\subsection{Attribute-based sampling and analysis}

\figResultsAblation{tab:resablations}

\figDisentanglement{tab:disent}

A key feature of our discrete latent formulation, is that it provides for a simple and effective method for analysis and guided sampling. In this section, we introduce an approach to attribute-based analysis and conditional sampling, by aggregating statistics of how a set of image-specific attributes relate to individual elements in the codebook.  

Let $\{a_1, \ldots, a_L\}$ denote the of attributes that are used to describe an image, for the specific dataset on which our generator is trained. Each attribute $a_l$ can take a finite set of values. For instance, in case of a face dataset, an attribute can describe the existence of glasses, beard, lipstick, or the hair color. In order to perform conditional image generation given a specified set of attributes, we need to estimate the conditional latent distribution $p(k|a_1, \ldots, a_L)$. We assume the positions to be conditionally independent $p(k|a_1, \ldots, a_L) = \prod_i p(k_i | a_1, \ldots, a_L)$. We then obtain,
\begin{align}
\begin{split}
\label{eq:var-given-att}
    p(k_i | a_1, \ldots, a_L) &= \frac{p(a_1, \ldots, a_L | k_i) p(k_i)}{\sum_{k_i}p(a_1, \ldots, a_L | k_i) p(k_i)} = \\
    \frac{\prod_{l}p(a_l | k_i)}{\sum_{k_i}\prod_{l}p(a_l | k_i)}
\end{split}\vspace{-2mm}
\end{align}
The first equality is the application of Bayes' rule. In the second equality, we use that $p(k_i) = \frac{1}{n_v}$ is uniform and assume the attributes to be conditionally independent given the variant $k_i$. The latter assumption is motivated by the high degree of disentanglement that we observe across variants and positions. Further, note that this conditional independence assumptions by no means imply that the generated attributes themselves are independent. In fact, as observed in our experiments, our approach captures the strong correlations that exist between certain attributes, such as `male' and `beard' (see our genome analysis and Figure \ref{fig:analysis}). 

Eq.~\ref{eq:var-given-att} shows that the conditional distribution of the latents are fully given by the marginal attribute distribution for a given embedding variant $p(a_l | k_i)$. We estimate the latter by aggregating statistics over generated image samples as
\begin{align}
\label{eq:att-given-var}
\begin{split}
    p(a_l | k_i=j) = \sum_k p(a_l | G(V_k)) p(k | k_i=j) \approx \\
    \frac{\sum_{k \in S: k_i = j} p(a_l|G(V_k))}{\sum_{k \in S: k_i = j} 1}
\end{split}\vspace{-2mm}
\end{align}
Here, $p(a_l | G(V_k))$ is the attribute distribution of the generated image $G(V_k)$, which we estimate with a pre-trained image classifier.
In the first equality, we marginalize over all possible latent vectors $k$. However, as this is intractable, we approximate the expectation value through Monte-Carlo sampling. Specifically, we pre-generate a set of images $\{G(V_k) : k \in S\}$, where the latents in $S$ are sampled from $p(k)$. 
We can efficiently re-use the same set of images, generated from $S$, when computing \eqref{eq:att-given-var} for all variants $k_i$ and attributes $l$. 

To further increase the likelihood of sampling codebook entries with high probability of the conditioned attribute class, we scale the estimated statistics with a temperature parameter $p(a_l | k_i)^{\frac{1}{T}}$ when employed in \eqref{eq:var-given-att}. This serves to increase the class consistency of the conditional sampling in our experiments. 

%% file: source/experiments.tex
\section{Experiments}\label{sec:experiments}

\figVectorQuantization{tab:vq}

\figAnalysis{fig:analysis}

\textbf{Implementation}
Our method, StyleGenes, is written in Pytorch~\cite{pytorch}. We incorporate our sampling approach into two baseline models: (1) StyleGAN2~\cite{Karras2019stylegan2} as provided in the 
StyleGAN3~\cite{stylegan3Karras2021} codebase and ProjectedGANs~\cite{Sauer2021ProjectedGC} using the FastGAN~\cite{liu2021towards} generator. 
For all datasets, we train all our models and baselines \textit{unconditionally} using 4 GPUs following the default configuration as described in each project's code repository~\cite{stylegan3Karras2021,Sauer2021ProjectedGC}. For small datases Metfaces~\cite{Karras2020ada} and AFHQ~\cite{choi2020starganv2} we use adaptive discriminator augmentation~\cite{Karras2020ada}.
For our StyleGAN2 experiments, we train until the discriminator has seen 10 million images of resolution $256 \times 256$.
For ProjectedGAN, we train for their reported number of iterations to reach state-of-the-art results, rounded up to the next million: 8M images for FFHQ\cite{Karras_2019_CVPR} and 2M images for the other datasets. 

\textbf{Datasets}
We investigate the performance of our network using the \textit{Fr\'echet Inception Distance (FID)}~\cite{NIPS2017_7240}, on widely used datasets for unsupervised image generation:\\
\textbf{FFHQ}~\cite{Karras_2019_CVPR}, is a collection of 70,000 face images scraped from flickr.com. The images were centered around the eyes and the mouth of the individual, offering strong position priors. The people depicted in the images come from a diverse background, age and poses.\\ 
\textbf{MetFaces}~\cite{Karras2020ada} is a dataset of image crops from art pieces of the Metropolitan Museum of Art Collection.
Similarly to FFHQ the crops are centered around human faces. 
The dataset contain 1336 images in total. The images are under CC0 license by the Metropolitan Museum of Art. 
\textbf{AFHQ}~\cite{choi2020starganv2} is a collection of 15.000 images of animal faces divided equally into three categories: cat, dog and wildlife. However, in this work we do not use the labels for conditional generation.\\ 
\textbf{LSUN Church \& Bedroom}~\cite{yu15lsun}. We are using two subsets of the LSUN dataset \emph{Church} and \emph{Bedroom}, where they contain diverse outdoor and indoor scenes respectively. We use the full LSUN Church dataset of 126,227 images and a subset of the bedroom scenes comprised of 121,000 images. 

\subsection{Unconditional Generation}

\figConditional{fig:conditional}

In Table~\ref{tab:resablations}-A we see the performance of established baselines~\cite{Karras2019stylegan2, Sauer2021ProjectedGC} using StyleGenes.
We compare with the continuous approach by training our baselines~\cite{Karras2020ada,Sauer2021ProjectedGC} with the same hyperparameters and number of images.  
Our proposed discrete method produces similar results with StyleGAN's StyleMapping approach, and improves ProjectedGAN when it replaces Gaussian sampling. 
\textcolor{\checkColor}{Note, that StyleGAN2 failed to converge in our Beds experiments.}

In our ablation study (Table~\ref{tab:resablations}-B), we analyze the effect of the different Genome configurations to the perceptual performance of the network, while keeping the size of the resulting latent vector fixed at $d=512$. 

Increasing the number of different variants for each  gene increases the number of parameters: $n_{v} * d$.
Doing so yields better FID scores.

Note that every experiment that is in the same row in Table~\ref{tab:resablations}-B is using the same number of parameters.
When we change the number of genes $n_g$, we also change the size of each sub-vector/variant $v_i^{k_i}$, such as $n_g * len({v_i}^{k_i}) = d$. 
Thus, by increasing the number of genes we do not increase the memory footprint, but 
the number of different images the genome can represent is also increasing, per Eq.~\eqref{eq:combo}.
This also leads to a decrease of FID.
In Table~\ref{tab:resablations}-right we can observe that for all three datasets, for the smallest gene length, going from variants' number of $2048$ to $256$ leads to a minor deterioration of performance. However, the memory footprint of the genome is decreased 8-fold. 

To summarize, we observe that we can increase the performance of our network by either increase its parameters by increasing the number of variants, or by dividing it up to more genes without an increase of memory.

\textcolor{\checkColor}{Additionally, in Table \ref{tab:vq} we provide a comparison with methods that are using vector quantization~\cite{VQGANEsser2021TamingTF,vitvqgan2021}.}

\subsection{Analyzing the Codebook}

In this section we exploit the discrete nature of our approach to analyze the association of the latent codes and their corresponding attributes. We show how our method improves disentanglement and how to harness our analysis to use our unconditionally trained model for conditional generation. Lastly, we show how to do interpolation and projection in our setting. 

\textbf{Associating variants with attributes} 
As described in Sec. \ref{sec:method} we run a Monte Carlo experiment to estimate the probability $p(a_l | k_i=j)$ of the variant $j$ at position $i$ resulting to the attribute $a_l$ in the output image. 
We randomly sample $500,000$ gene sequences from our FFHQ model and generate their corresponding images.
We pass each of these images through 40 pretrained CelebA classifiers~\cite{celeba2015liu}.
We used the weights provided by the original StyleGAN~\cite{Karras_2019_CVPR} repository, and the code provided by StyleSpace~\cite{stylespaceWu2021} to extract the logits for every image.
For instance, let $i=15$ and $j=217$, and $a_l$ a facial attribute, such as \emph{black hair}. 
We estimate $p(\textit{blackhair}| k_{15}=217)$ by averaging the outputs of the \emph{black hair} classifier for \emph{all} style codes containing variant $217$ in their $17^\textit{th}$ position. 
We repeat the process for each variant and attribute to get the marginal attribute distribution for a given genome variant.

\textbf{Which genes are responsible for each attribute?}
We want to test if, like its biological inspiration, our genome has specific genes that control the expression of certain attributes, such as hair color.
We would like to measure the impact that changing a gene has to a certain trait of the output image.
We hypothesize that if a gene controls an attribute, it will exhibit high variance in its expected values and have extreme values towards both ends.
We quantify a gene's importance by calculating the \emph{mean absolute standard score} for each gene position: the absolute distance in terms of standard deviations that the gene variants have on average with the codebook's mean expected value for the particular attribute:
$s_{l}^{i} = \Sigma_{j}\frac{\lvert p(a_l| k_i=j) - \mu_{p(a_l| k_i)} \rvert}{\sigma_{p(a_l| k_i)}}$. 


In Figure~\ref{fig:analysis} we see the score for a gene in a specific position.
The genes are placed circularly around the plot.
Each color represents one of the 40 attributes.
Most genes do not significantly affect any of the attributes, instead controlling local image details. 
For each attribute, only a handful of genes have high standard scores.
On the right side of Figure~\ref{fig:analysis}, we see how changing the variants of specific genes alters the output image.
We sample a gene sequence and start substituting the variant of one gene at random. 
Manipulating the genes that exhibit high scores in the polar plot, such as genes 66, 85, and 38, leads to visible changes in the image.  
However, most genes do not exhibit large scores for any of the attributes.
Changing these genes' variants results in barely noticeable changes.

\figInterpolation{fig:interpolation}
\textbf{Conditional Generation}
In the previous steps we acquired the marginal attribute distribution for a given variant. We use this information to conditionally generate an image with a desired attribute $a_l$. In Tab.~\ref{fig:conditional} we can see samples produced by our method. To generate unconditionally we sample the variant for each gene position uniformly. However, as described in Section \ref{sec:method} we can now infer the conditional latent distribution $p(k|{a_l})$ and use it to sample the variants instead. In Figure \ref{fig:conditional} we can see the results of our conditional sampling. Decreasing the temperature $t$ increases the likelihood of the presence of the desired attribute, however, can also limit the variability of the conditional outputs, as it is outlined in increased FID scores in Tab.~\ref{fig:conditional}. 

To gauge the ability of our method to generate conditionally, we train a StyleGAN2 model with pseudo-labels from the CelebA classifiers. Training conditionally, limits the model's generation to a fixed set of attributes. Additionally, every attribute needs to have a value, 0 or 1, in order to generate a sample. For our experiment, we sample all other attributes than the one we aim to generate either \emph{randomly} or by using their co-occurance \emph{frequency} in FFHQ.

In Tab.~\ref{fig:conditional}, we find we compare similarly to our baseline. However, we are not limited  by a predefined number of attributes and can be extend to more without training. Moreover, we can use the temperature value to control the trade-off between variability and accuracy.  
Lastly, training conditionally with a small dataset can lead to poor performance and mode collapse~\cite{shahbazi2022collapse}. With our method we are conditionally sampling gene variants from our unconditionally trained model and, thus, we do not face the same problem.

\textbf{StyleGenome and Disentanglement}
The StyleGAN's~\cite{Karras_2019_CVPR} motivation to design the StyleMapping Network to make the sampling density determined by the mapping and not to be limited to any fixed distribution;
they aimed for the resulting space W to be more disentangled.
We explore how disentangled our \emph{StyleGenes} are compared to the output space of StyleMapping.
We train a Multi-Layer Perceptron to predict the presence of an attributed in an image from its latent code.
We randomly sample 50.000 codes from each of the two representations.
Then we extract the fake images' attributes using the pretrained CelebA classifiers, and appropriately prepared the train/val/test subsets.
We find that StyleGenes outperforms the StyleMapping's accuracy on every attribute we tested, with a ~10\% average increase, as shown in Tab.~\ref{tab:disent}.
In Figure~\ref{fig:analysis}, we see that certain genes affect a group of pertinent attributes.
During training, each variant is sampled independently from the rest. We hypothesize that this pushes the variants to be semantically self-contained compared to the StyleMapping approach, which maps an undivided vector to the W space.


\textbf{Interpolation}.
During training we sample the latent codes from our discrete codebook, but their values lie on $\mathbb{R}^d$.
We want to gauge whether the learned genome comprises samples that lie on a smooth surface. 
We sample two codes and interpolate between them. In Figure~\ref{fig:interpolation} we can see interpolation results for all three datasets.
The transition is smooth and the subsequent samples are semantically coherent and realistic.
By optimizing on discrete samples we are able to learn a continuous distribution.

\textbf{Adding real images in the codebook}.
We extend the Pivotal Tuning Inversion~\cite{roich2021pivotal} to project real images into our codebook in Fig.~\ref{fig:interpolation}. We start by concurrently optimizing a set of vectors in the underlying continuous space to produce the images we aim to invert. Then, we find the indices of the nearest-neighbor variant for each gene in the codebook. We train both the generator and the codebook to recreate the images, based only on these indices. We substitute PTI's locality regularization with our codebook-perseverance regularization: we push randomly sampled codes to keep their syntheses unchanged via an LPIPS~\cite{zhang2018perceptual} loss. We find this step important to retain the perceptual quality of the genome.



%% file: source/conclusion.tex
\section{Conclusion}

In this work we introduce \emph{StyleGenes}.
Inspired by how information is encoded in the DNA by only four basic building blocks, we design a discrete sampling approach for GANs.
We define our StyleGenome, an ordered collection of gene variants. We uniformly sample a variant for each gene to form a sequence. Its concatenation is the style code used by the generator to synthesize an image.
Our discrete sampling technique achieves an FID score on par with its continuous counterpart, while enabling an intuitive way to analyze the latent code. 
We use pretrained classifiers to aggregate attribute statistics, enabling attribute-based analysis. 
Our analysis enables conditional sampling out of our unconditionally trained model.
Lastly, we show that we can generate samples between the genome's discrete elements, indicating that the samples are on a smooth style surface, and devise an approach to incorporate real images in our genome. 



%% file: source/limitations.tex
\section{Ethical Discussion and Limitations}

Our methodology permits exploiting the discrete latent code of an image generator trained in an unsupervised manner to generate conditionally. 
However, it comes with some limitations.
First, while GANs aim to mimic the distribution of real images there is a gap between the real and fake distribution~\cite{precisionrecall}.
We apply a classifier trained on real images to infer the labels of fake images.
The classifiers are limited by the entanglements of attributes in the training dataset. 
This issue is also discussed in StyleSpace~\cite{stylespaceWu2021}, where they note that the classifier may fail to predict a lipstick on a male face.
We hypothesize that this effect is further intensified because, while FFHQ offers significant diversity in ``age, ethnicity and accessories'', CelebA contains celebrity faces and thus is more limited in those factors.
Lastly, the biases of the labelers of the images can be propagated to the conditional image generation 
results (\emph{attractive} is among CelebA's attributes).
Therefore, one should be wary to apply this approach to a real life application.

Please note that in this work we train our model using portraits of real people, using the Flickr-Faces-HQ dataset~\cite{Karras_2019_CVPR}. As described in \emph{https://github.com/NVlabs/ffhq-dataset}, the images were collected to adhere to privacy rules and were filtered to only include samples intented for redistribution. Moreover, if an individual identifies themselves in the dataset, they can request the removal of their face from the collection.

%% file: main.bbl
\begin{thebibliography}{10}\itemsep=-1pt

\bibitem{embedstylegan}
Rameen Abdal, Yipeng Qin, and Peter Wonka.
\newblock Image2stylegan: How to embed images into the stylegan latent space?
\newblock In {\em 2019 IEEE/CVF International Conference on Computer Vision
  (ICCV)}, pages 4431--4440, 2019.

\bibitem{styleflow}
Rameen Abdal, Peihao Zhu, Niloy~J. Mitra, and Peter Wonka.
\newblock Styleflow: Attribute-conditioned exploration of stylegan-generated
  images using conditional continuous normalizing flows.
\newblock {\em ACM Trans. Graph.}, 40(3), May 2021.

\bibitem{brock2018large}
Andrew Brock, Jeff Donahue, and Karen Simonyan.
\newblock Large scale {GAN} training for high fidelity natural image synthesis.
\newblock In {\em Proceedings of the International Conference on Learning
  Representations (ICLR)}, 2019.

\bibitem{chanmonteiro2020pi-GAN}
Eric Chan, Marco Monteiro, Petr Kellnhofer, Jiajun Wu, and Gordon Wetzstein.
\newblock pi-gan: Periodic implicit generative adversarial networks for
  3d-aware image synthesis.
\newblock In {\em arXiv}, 2020.

\bibitem{eg3d}
Eric~R. Chan, Connor~Z. Lin, Matthew~A. Chan, Koki Nagano, Boxiao Pan,
  Shalini~De Mello, Orazio Gallo, Leonidas Guibas, Jonathan Tremblay, Sameh
  Khamis, Tero Karras, and Gordon Wetzstein.
\newblock Efficient geometry-aware {3D} generative adversarial networks.
\newblock In {\em CVPR}, 2022.

\bibitem{10.5555/3157096.3157340}
Xi Chen, Yan Duan, Rein Houthooft, John Schulman, Ilya Sutskever, and Pieter
  Abbeel.
\newblock Infogan: Interpretable representation learning by information
  maximizing generative adversarial nets.
\newblock In {\em Proceedings of the 30th International Conference on Neural
  Information Processing Systems}, NIPS'16, page 2180–2188, Red Hook, NY,
  USA, 2016. Curran Associates Inc.

\bibitem{choi2020starganv2}
Yunjey Choi, Youngjung Uh, Jaejun Yoo, and Jung-Woo Ha.
\newblock Stargan v2: Diverse image synthesis for multiple domains.
\newblock In {\em Proceedings of the IEEE Conference on Computer Vision and
  Pattern Recognition}, 2020.

\bibitem{Collins2020EditingIS}
Edo Collins, Raja Bala, Bob Price, and Sabine S{\"u}sstrunk.
\newblock Editing in style: Uncovering the local semantics of gans.
\newblock {\em 2020 IEEE/CVF Conference on Computer Vision and Pattern
  Recognition (CVPR)}, pages 5770--5779, 2020.

\bibitem{VQGANEsser2021TamingTF}
Patrick Esser, Robin Rombach, and Bj{\"o}rn Ommer.
\newblock Taming transformers for high-resolution image synthesis.
\newblock {\em 2021 IEEE/CVF Conference on Computer Vision and Pattern
  Recognition (CVPR)}, pages 12868--12878, 2021.

\bibitem{goodfellowGAN}
Ian Goodfellow, Jean Pouget-Abadie, Mehdi Mirza, Bing Xu, David Warde-Farley,
  Sherjil Ozair, Aaron Courville, and Yoshua Bengio.
\newblock Generative adversarial nets.
\newblock In Z. Ghahramani, M. Welling, C. Cortes, N.~D. Lawrence, and K.~Q.
  Weinberger, editors, {\em Advances in Neural Information Processing Systems},
  pages 2672--2680, 2014.

\bibitem{NIPS2017_892c3b1c}
Ishaan Gulrajani, Faruk Ahmed, Martin Arjovsky, Vincent Dumoulin, and Aaron~C
  Courville.
\newblock Improved training of wasserstein gans.
\newblock In I. Guyon, U.~Von Luxburg, S. Bengio, H. Wallach, R. Fergus, S.
  Vishwanathan, and R. Garnett, editors, {\em Advances in Neural Information
  Processing Systems}, volume~30. Curran Associates, Inc., 2017.

\bibitem{improvedwgans}
Ishaan Gulrajani, Faruk Ahmed, Martin Arjovsky, Vincent Dumoulin, and Aaron~C
  Courville.
\newblock Improved training of wasserstein gans.
\newblock In I. Guyon, U.~Von Luxburg, S. Bengio, H. Wallach, R. Fergus, S.
  Vishwanathan, and R. Garnett, editors, {\em Advances in Neural Information
  Processing Systems}, volume~30. Curran Associates, Inc., 2017.

\bibitem{NIPS2017_7240}
Martin Heusel, Hubert Ramsauer, Thomas Unterthiner, Bernhard Nessler, and Sepp
  Hochreiter.
\newblock Gans trained by a two time-scale update rule converge to a local nash
  equilibrium.
\newblock In I. Guyon, U.~V. Luxburg, S. Bengio, H. Wallach, R. Fergus, S.
  Vishwanathan, and R. Garnett, editors, {\em Advances in Neural Information
  Processing Systems}, pages 6626--6637, 2017.

\bibitem{adain2017huang}
Xun Huang and Serge Belongie.
\newblock Arbitrary style transfer in real-time with adaptive instance
  normalization, 2017.

\bibitem{jahanian2020steerability}
Ali Jahanian, Lucy Chai, and Phillip Isola.
\newblock On the "steerability" of generative adversarial networks, 2020.

\bibitem{karras2018progressive}
Tero Karras, Timo Aila, Samuli Laine, and Jaakko Lehtinen.
\newblock Progressive growing of {GAN}s for improved quality, stability, and
  variation.
\newblock In {\em Proceedings of the International Conference on Learning
  Representations (ICLR)}, 2018.

\bibitem{Karras2020ada}
Tero Karras, Miika Aittala, Janne Hellsten, Samuli Laine, Jaakko Lehtinen, and
  Timo Aila.
\newblock Training generative adversarial networks with limited data.
\newblock In {\em Proc. NeurIPS}, 2020.

\bibitem{stylegan3Karras2021}
Tero Karras, Miika Aittala, Samuli Laine, Erik H\"ark\"onen, Janne Hellsten,
  Jaakko Lehtinen, and Timo Aila.
\newblock Alias-free generative adversarial networks.
\newblock In {\em Proc. NeurIPS}, 2021.

\bibitem{Karras_2019_CVPR}
Tero Karras, Samuli Laine, and Timo Aila.
\newblock A style-based generator architecture for generative adversarial
  networks.
\newblock In {\em Proceedings of the IEEE Conference on Computer Vision and
  Pattern Recognition (CVPR)}, 2019.

\bibitem{Karras2019stylegan2}
Tero Karras, Samuli Laine, Miika Aittala, Janne Hellsten, Jaakko Lehtinen, and
  Timo Aila.
\newblock Analyzing and improving the image quality of {StyleGAN}.
\newblock In {\em Proc. CVPR}, 2020.

\bibitem{precisionrecall}
Tuomas Kynkäänniemi, Tero Karras, Samuli Laine, Jaakko Lehtinen, and Timo
  Aila.
\newblock Improved precision and recall metric for assessing generative models.
\newblock {\em CoRR}, abs/1904.06991, 2019.

\bibitem{explaining_in_style}
Oran Lang, Yossi Gandelsman, Michal Yarom, Yoav Wald, Gal Elidan, Avinatan
  Hassidim, William~T. Freeman, Phillip Isola, Amir Globerson, Michal Irani,
  and Inbar Mosseri.
\newblock Explaining in style: Training a gan to explain a classifier in
  stylespace.
\newblock {\em arXiv preprint arXiv:2104.13369}, 2021.

\bibitem{lee2022autoregressive}
Doyup Lee, Chiheon Kim, Saehoon Kim, Minsu Cho, and Wook-Shin Han.
\newblock Autoregressive image generation using residual quantization.
\newblock {\em arXiv preprint arXiv:2203.01941}, 2022.

\bibitem{liu2021towards}
Bingchen Liu, Yizhe Zhu, Kunpeng Song, and Ahmed Elgammal.
\newblock Towards faster and stabilized {\{}gan{\}} training for high-fidelity
  few-shot image synthesis.
\newblock In {\em International Conference on Learning Representations}, 2021.

\bibitem{celeba2015liu}
Ziwei Liu, Ping Luo, Xiaogang Wang, and Xiaoou Tang.
\newblock Deep learning face attributes in the wild.
\newblock In {\em Proceedings of International Conference on Computer Vision
  (ICCV)}, December 2015.

\bibitem{styleEMB_neurips2021}
Weili Nie, Arash Vahdat, and Anima Anandkumar.
\newblock Controllable and compositional generation with latent-space
  energy-based models.
\newblock In M. Ranzato, A. Beygelzimer, Y. Dauphin, P.S. Liang, and J.~Wortman
  Vaughan, editors, {\em Advances in Neural Information Processing Systems},
  volume~34, pages 13497--13510. Curran Associates, Inc., 2021.

\bibitem{ntavelis2020sesame}
Evangelos Ntavelis, Andr{\'{e}}s Romero, Iason Kastanis, Luc {Van Gool}, and
  Radu Timofte.
\newblock {SESAME: Semantic Editing of Scenes by Adding, Manipulating or
  Erasing Objects}.
\newblock In Andrea Vedaldi, Horst Bischof, Thomas Brox, and Jan-Michael Frahm,
  editors, {\em Computer Vision -- ECCV 2020}, pages 394--411, Cham, 2020.
  Springer International Publishing.

\bibitem{ntavelis2022arbitrary}
Evangelos Ntavelis, Mohamad Shahbazi, Iason Kastanis, Radu Timofte, Martin
  Danelljan, and Luc Van~Gool.
\newblock Arbitrary-scale image synthesis.
\newblock In {\em 2022 {IEEE} Conference on Computer Vision and Pattern
  Recognition, {CVPR} 2022}, 2022.

\bibitem{park2019SPADE}
Taesung Park, Ming-Yu Liu, Ting-Chun Wang, and Jun-Yan Zhu.
\newblock Semantic image synthesis with spatially-adaptive normalization.
\newblock In {\em Proceedings of the IEEE Conference on Computer Vision and
  Pattern Recognition (CVPR)}, 2019.

\bibitem{pytorch}
Adam Paszke, Sam Gross, Francisco Massa, Adam Lerer, James Bradbury, Gregory
  Chanan, Trevor Killeen, Zeming Lin, Natalia Gimelshein, Luca Antiga, Alban
  Desmaison, Andreas Kopf, Edward Yang, Zachary DeVito, Martin Raison, Alykhan
  Tejani, Sasank Chilamkurthy, Benoit Steiner, Lu Fang, Junjie Bai, and Soumith
  Chintala.
\newblock Pytorch: An imperative style, high-performance deep learning library.
\newblock In H. Wallach, H. Larochelle, A. Beygelzimer, F. d\textquotesingle
  Alch\'{e}-Buc, E. Fox, and R. Garnett, editors, {\em Advances in Neural
  Information Processing Systems 32}, pages 8024--8035. Curran Associates,
  Inc., 2019.

\bibitem{patashnik2021styleclip}
Or Patashnik, Zongze Wu, Eli Shechtman, Daniel Cohen-Or, and Dani Lischinski.
\newblock Styleclip: Text-driven manipulation of stylegan imagery, 2021.

\bibitem{Radford2015UnsupervisedRL}
Alec Radford, Luke Metz, and Soumith Chintala.
\newblock Unsupervised representation learning with deep convolutional
  generative adversarial networks.
\newblock {\em CoRR}, abs/1511.06434, 2015.

\bibitem{VQVAE2Razavi2019GeneratingDH}
Ali Razavi, A{\"a}ron van~den Oord, and Oriol Vinyals.
\newblock Generating diverse high-fidelity images with vq-vae-2.
\newblock {\em ArXiv}, abs/1906.00446, 2019.

\bibitem{pmlr-v48-reed16}
Scott Reed, Zeynep Akata, Xinchen Yan, Lajanugen Logeswaran, Bernt Schiele, and
  Honglak Lee.
\newblock Generative adversarial text to image synthesis.
\newblock In Maria~Florina Balcan and Kilian~Q. Weinberger, editors, {\em
  Proceedings of The 33rd International Conference on Machine Learning},
  volume~48 of {\em Proceedings of Machine Learning Research}, pages
  1060--1069, New York, New York, USA, 20--22 Jun 2016. PMLR.

\bibitem{roich2021pivotal}
Daniel Roich, Ron Mokady, Amit~H Bermano, and Daniel Cohen-Or.
\newblock Pivotal tuning for latent-based editing of real images.
\newblock {\em ACM Trans. Graph.}, 2021.

\bibitem{Sauer2021ProjectedGC}
Axel Sauer, Kashyap Chitta, Jens Muller, and Andreas Geiger.
\newblock Projected gans converge faster.
\newblock In {\em NeurIPS}, 2021.

\bibitem{Sauer2021ARXIV}
Axel Sauer, Katja Schwarz, and Andreas Geiger.
\newblock Stylegan-xl: Scaling stylegan to large diverse datasets.
\newblock volume abs/2201.00273, 2022.

\bibitem{oasis}
Edgar Sch{\"o}nfeld, Vadim Sushko, Dan Zhang, Juergen Gall, Bernt Schiele, and
  Anna Khoreva.
\newblock You only need adversarial supervision for semantic image synthesis.
\newblock In {\em International Conference on Learning Representations}, 2021.

\bibitem{shahbazi2022collapse}
Mohamad Shahbazi, Martin Danelljan, Danda~Pani Paudel, and Luc~Van Gool.
\newblock Collapse by conditioning: Training class-conditional {GAN}s with
  limited data.
\newblock In {\em International Conference on Learning Representations}, 2022.

\bibitem{shahbazi2021cGANTransfer}
Mohamad Shahbazi, Zhiwu Huang, Danda~Pani Paudel, Ajad Chhatkuli, and Luc
  Van~Gool.
\newblock Efficient conditional gan transfer with knowledge propagation across
  classes.
\newblock In {\em 2021 {IEEE} Conference on Computer Vision and Pattern
  Recognition, {CVPR} 2021}, 2021.

\bibitem{VQVAEOord2017NeuralDR}
A{\"a}ron van~den Oord, Oriol Vinyals, and Koray Kavukcuoglu.
\newblock Neural discrete representation learning.
\newblock In {\em NIPS}, 2017.

\bibitem{voynov2020unsupervised}
Andrey Voynov and Artem Babenko.
\newblock Unsupervised discovery of interpretable directions in the gan latent
  space.
\newblock In {\em International Conference on Machine Learning}, pages
  9786--9796. PMLR, 2020.

\bibitem{wang2018pix2pixHD}
Ting-Chun Wang, Ming-Yu Liu, Jun-Yan Zhu, Andrew Tao, Jan Kautz, and Bryan
  Catanzaro.
\newblock High-resolution image synthesis and semantic manipulation with
  conditional gans.
\newblock In {\em Proceedings of the IEEE Conference on Computer Vision and
  Pattern Recognition (CVPR)}, 2018.

\bibitem{samplingGANs}
Tom White.
\newblock Sampling generative networks, 2016.

\bibitem{stylespaceWu2021}
Zongze Wu, Dani Lischinski, and Eli Shechtman.
\newblock Stylespace analysis: Disentangled controls for stylegan image
  generation.
\newblock In {\em Proceedings of the IEEE/CVF Conference on Computer Vision and
  Pattern Recognition (CVPR)}, pages 12863--12872, June 2021.

\bibitem{dnaganxiao2018}
Taihong Xiao, Jiapeng Hong, and Jinwen Ma.
\newblock Dna-gan: Learning disentangled representations from multi-attribute
  images.
\newblock {\em International Conference on Learning Representations, Workshop},
  2018.

\bibitem{ElegantXiao2018}
Taihong Xiao, Jiapeng Hong, and Jinwen Ma.
\newblock Elegant: Exchanging latent encodings with gan for transferring
  multiple face attributes.
\newblock In {\em Proceedings of the European Conference on Computer Vision
  (ECCV)}, pages 172--187, September 2018.

\bibitem{attngan}
Tao Xu, Pengchuan Zhang, Qiuyuan Huang, Han Zhang, Zhe Gan, Xiaolei Huang, and
  Xiaodong He.
\newblock Attngan: Fine-grained text to image generation with attentional
  generative adversarial networks.
\newblock {\em arXiv preprint arXiv:1711.10485}, 2017.

\bibitem{yang2019semantic}
Ceyuan Yang, Yujun Shen, and Bolei Zhou.
\newblock Semantic hierarchy emerges in deep generative representations for
  scene synthesis.
\newblock {\em International Journal of Computer Vision}, 2020.

\bibitem{yu15lsun}
Fisher Yu, Yinda Zhang, Shuran Song, Ari Seff, and Jianxiong Xiao.
\newblock Lsun: Construction of a large-scale image dataset using deep learning
  with humans in the loop.
\newblock {\em arXiv preprint arXiv:1506.03365}, 2015.

\bibitem{vitvqgan2021}
Jiahui Yu, Xin Li, Jing~Yu Koh, Han Zhang, Ruoming Pang, James Qin, Alexander
  Ku, Yuanzhong Xu, Jason Baldridge, and Yonghui Wu.
\newblock Vector-quantized image modeling with improved vqgan, 2021.

\bibitem{zhang2018perceptual}
Richard Zhang, Phillip Isola, Alexei~A Efros, Eli Shechtman, and Oliver Wang.
\newblock The unreasonable effectiveness of deep features as a perceptual
  metric.
\newblock In {\em CVPR}, 2018.

\bibitem{zhang21}
Yuxuan Zhang, Huan Ling, Jun Gao, Kangxue Yin, Jean-Francois Lafleche, Adela
  Barriuso, Antonio Torralba, and Sanja Fidler.
\newblock Datasetgan: Efficient labeled data factory with minimal human effort.
\newblock In {\em CVPR}, 2021.

\bibitem{pmlr-v119-zhao20d}
Yang Zhao, Chunyuan Li, Ping Yu, Jianfeng Gao, and Changyou Chen.
\newblock Feature quantization improves {GAN} training.
\newblock In Hal~Daumé III and Aarti Singh, editors, {\em Proceedings of the
  37th International Conference on Machine Learning}, volume 119 of {\em
  Proceedings of Machine Learning Research}, pages 11376--11386. PMLR, 13--18
  Jul 2020.

\bibitem{Zhu2020SEANIS}
Peihao Zhu, Rameen Abdal, Yipeng Qin, and Peter Wonka.
\newblock Sean: Image synthesis with semantic region-adaptive normalization.
\newblock {\em 2020 IEEE/CVF Conference on Computer Vision and Pattern
  Recognition (CVPR)}, pages 5103--5112, 2020.

\end{thebibliography}
